\documentclass{article}

\usepackage{amsmath,amsfonts,bm}

\def\eqref#1{equation~\ref{#1}}

\def\1{\bm{1}}

\def\vx{{\bm{x}}}
\def\vy{{\bm{y}}}

\DeclareMathAlphabet{\mathsfit}{\encodingdefault}{\sfdefault}{m}{sl}
\SetMathAlphabet{\mathsfit}{bold}{\encodingdefault}{\sfdefault}{bx}{n}

\def\gB{{\mathcal{B}}}

\def\gD{{\mathcal{D}}}

\def\gS{{\mathcal{S}}}

\def\sR{{\mathbb{R}}}

\DeclareMathOperator*{\argmin}{arg\,min}

\newcommand{\metric}[2]{#1$_{\textcolor{gray}{\scriptsize\pm#2}}$}
\usepackage{xcolor}
\definecolor{mainblue}{RGB}{21, 61, 131}     
\definecolor{lightblue}{RGB}{98, 139, 190}  

\newcommand{\metricbl}[2]{\textcolor{mainblue}{#1}$_{\textcolor{lightblue}{\scriptsize\pm#2}}$}
\newcommand{\metricblbold}[1]{\textbf{\textcolor{mainblue}{#1}}}

\usepackage[preprint]{neurips_2025}

\usepackage[utf8]{inputenc} 
\usepackage[T1]{fontenc}    
\usepackage{url}            
\usepackage{booktabs}       
\usepackage{amsfonts}       
\usepackage{nicefrac}       
\usepackage{microtype}      
\usepackage{xcolor}         
\usepackage{graphicx}
\usepackage{wrapfig}
\usepackage{xcolor}
\definecolor{mycitecolor}{HTML}{3c6db0}
\definecolor{mylinkcolor}{HTML}{bc3c53}
\usepackage{multirow}
\usepackage{array}  
\usepackage[colorlinks=true, citecolor=mycitecolor, linkcolor=mylinkcolor, urlcolor=magenta]{hyperref}
\usepackage{algorithm}
\usepackage{algpseudocode}

\usepackage{colortbl} 
\usepackage{tikz}
\usetikzlibrary{matrix}
\usepackage{subcaption}
\usepackage{caption}
\usepackage{geometry}
\title{Beyond Modality Collapse: Representations Blending for Multimodal Dataset Distillation}

\author{Xin Zhang\textsuperscript{1,2} \quad
Ziruo Zhang\textsuperscript{3} \quad
Jiawei Du\textsuperscript{1,2}\quad
Zuozhu Liu\textsuperscript{4}\quad 
Joey Tianyi Zhou\textsuperscript{1,2}\\
\textsuperscript{1}{\small Centre for Frontier AI Research, Agency for Science, Technology and Research, Singapore} \\
\textsuperscript{2}{\small Institute of High Performance Computing, Agency for Science, Technology and Research, Singapore}\\
\textsuperscript{3}{\small National University of Singapore, Singapore} \quad 
\textsuperscript{4}{\small Zhejiang University, China} \\
\texttt{\small \{zhangx7, dujw, Joey\_Zhou\}@cfar.a\-star.edu.sg} \\
\texttt{\small ziruo.z@u.nus.edu} \quad  \texttt{\small zuozhuliu@intl.zju.edu.cn}
}
\begin{document}

\maketitle
\begin{abstract}
Multimodal Dataset Distillation (MDD) seeks to condense large-scale image-text datasets into compact surrogates while retaining their effectiveness for cross-modal learning. Despite recent progress, existing MDD approaches often suffer from \textit{\textbf{Modality Collapse}}, characterized by over-concentrated intra-modal representations and enlarged distributional gap across modalities. In this paper, at the first time, we identify this issue as stemming from a fundamental conflict between the over-compression behavior inherent in dataset distillation and the cross-modal supervision imposed by contrastive objectives. To alleviate modality collapse, we introduce \textbf{RepBlend}, a novel MDD framework that weakens overdominant cross-modal supervision via representation blending, thereby significantly enhancing intra-modal diversity. Additionally, we observe that current MDD methods impose asymmetric supervision across modalities, resulting in biased optimization. To address this, we propose symmetric projection trajectory matching, which synchronizes the optimization dynamics using modality-specific projection heads, thereby promoting balanced supervision and enhancing cross-modal alignment.
Experiments on Flickr-30K and MS-COCO show that RepBlend consistently outperforms prior state-of-the-art MDD methods, achieving significant gains in retrieval performance (e.g., +9.4 IR@10, +6.3 TR@10 under the 100-pair setting) and offering up to 6.7$\times$ distillation speedup.
\end{abstract}
\section{Introduction} 
\begin{figure}[h]
    \centering
    \includegraphics[width=0.75\linewidth]{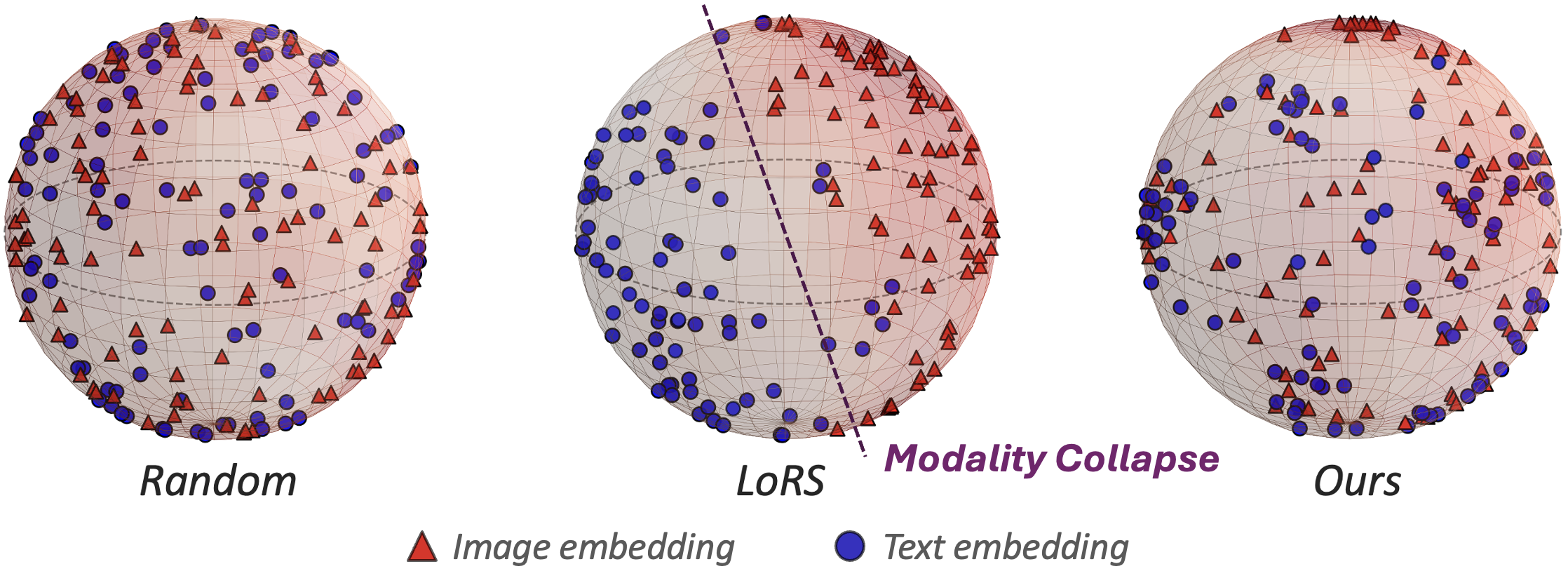}
    \caption{Multimodal embedding distributions across various distillation methods. We extract image and text embeddings from a finetuned CLIP~\cite{radford21a} and project them into a shared representation space using DOSNES~\cite{lu2019doubly}. Red triangles and blue circles denote image and text embeddings, respectively. \textbf{Left}: Embeddings from randomly sampled data in the original dataset exhibit a well-spread and modality-aligned distribution. \textbf{Middle}: The distilled dataset generated by a sota MDD method (LoRS~\cite{xu2024lors}) leads to \textit{Modality Collapse}, where image and text embeddings are poorly aligned and concentrated in distinct regions. \textbf{Right}: Our method effectively mitigates modality collapse, yielding a distribution that better preserves cross-modal alignment and exhibits greater representational diversity.}
    \vspace{-1.5em}
    \label{fig:modal_collapse}
\end{figure}
The unprecedented expansion of large-scale datasets has catalyzed recent breakthroughs in deep learning~\cite{chowdhery2023palm, bommasani2021opportunities, abnar2022exploring}, but has also introduced considerable storage and computational overhead~\cite{hoffmann2022training, kang2024get}. Thus, reducing dataset size to streamline the development process has emerged as an important research focus. Among various solutions, Dataset Distillation (DD)~\cite{dd2018} has emerged as a compelling strategy, achieving high compression ratios by synthesizing a compact surrogate dataset that approximates the training efficacy of the original dataset. The effectiveness of DD has been demonstrated across various modalities, including images~\cite{mtt, sre}, text~\cite{lu2025unidetox, maekawa2025dilm}, videos~\cite{ding2025condensing, wang2024dancing}, and graphs~\cite{liu2023graph, zhang2024navigating}. These unimodal successes motivate its extension to increasingly prominent multimodal scenarios~\cite{radford21a,liu2023visual,peng2024grounding,chen2023shikra}.

The pioneering effort in multimodal dataset distillation (MDD) is MTT-VL~\cite{wu2024visionlanguage}, which first validates the feasibility of extending existing vanilla DD techniques to the image-text setting. Building on this baseline, LoRS~\cite{xu2024lors} further proposes to mine cross-modal similarity to calibrate the supervision from matched and mismatched pairs, thereby achieving better adaptation to high-variance image-text data. Despite achieving promising results, existing studies remain confined to the data structure level, without probing the underlying conflict between DD and contrastive learning. Specifically, to prevent significant performance deterioration, vanilla DD prioritizes capturing representative features under limited distillation budgets, often sacrificing diversity and distributional coverage~\cite{seqmatch,datm, dwa2024neurips}. While this compromise is tolerable in unimodal classification tasks, naively applying such strategies to multimodal contrastive learning, which places great importance on instance-level discriminability, inevitably leads to \textbf{\textit{Modality Collapse}}. As illustrated in \autoref{fig:modal_collapse} (middle), the distilled dataset exhibits pronounced intra-modality aggregation and inter-modality separation.

This modality collapse leads to two critical issues.  
First, \textit{it induces excessive intra-modal similarity}, where embeddings within each modality become increasingly concentrated as distillation progresses. This over-concentration gradually suppresses representational diversity, making semantically distinct instances harder to separate, and eroding the fine-grained discrimination ability within each modality.
Second, \textit{it widens the inter-modal gap}, resulting in a large divergence between the feature distributions of different modalities. Insufficient cross-modal interaction fragments the embedding spaces and weakens semantic alignment, compromising the correct matching of positive pairs and the separation of negative pairs across modalities.

Recognizing these limitations, we propose \textbf{RepBlend}, a novel framework for MDD aimed at alleviating modality collapse. First, we theoretically identify that the collapse is induced by the over-compression nature of DD, where optimization converges toward a small set of dominant features. Cross-modal contrastive supervision further reinforces this convergence, leading to intra-modal collapse.
To address this issue, RepBlend introduces Representation Blending within each modality to weaken the overly strong cross-modal supervision, thereby promoting intra-modal diversity. 

Furthermore, we observe that existing MDD approaches exhibit asymmetric supervision between modalities, with the image branch receiving significantly weaker update signals than the text branch. To address this, we propose Symmetric Projection Trajectory Matching, a mechanism that aligns the optimization trajectories of both projection heads, thereby enhancing cross-modal alignment and improving overall distillation efficiency.
Extensive evaluations on Flickr-30K and MS-COCO demonstrate that RepBlend consistently surpasses existing MDD methods. Notably, under the 100-pair setting on Flickr-30K, it achieves improvements of +9.4 in IR@10 and +6.3 in TR@10, along with a 6.7$\times$ distillation speedup over the state-of-the-art baseline. \textcolor{black}{Beyond these benchmarks, RepBlend also exhibits strong generalization to other multimodal scenarios, such as audio-text.}

\textcolor{black}{Our contributions are summarized as follows:
\begin{itemize}
\item  For the first time, we identify the modality collapse issue in current MDD solutions, where the distilled dataset exhibits high intra-modal similarity and a large inter-modal gap. Through theoretical analysis, we attribute this to a mutually reinforcing effect between the over-compression behavior of dataset distillation and the cross-modal supervision enforced by contrastive objectives.
\item We propose Representation Blending to mitigate modality collapse by weakening the overly strong cross-modal supervision and enhancing intra-modal representational diversity. Furthermore, we introduce Symmetric Projection Trajectory Matching to enable more balanced multimodal distillation, which not only strengthens cross-modal alignment but also improves overall distillation efficiency.
\end{itemize}}
\section{Preliminaries and Related Works}
Dataset Distillation (DD)~\cite{dd2018} aims to synthesize a compact surrogate dataset by emulating the key properties of the original large dataset. These properties include distributional characteristics, such as feature-level statistics~\cite{DM, wang2022cafe, Minmax2025wang} and batch normalization parameters~\cite{sre, shao2023generalized, dwa2024neurips}, and training dynamics, including gradient~\cite{dc2021, dsa2021} and optimization trajectories~\cite{mtt, tesla, seqmatch, datm, lee2024selmatch}. While DD achieves promising results on unimodal benchmarks, extending it to multimodal scenarios remains challenging due to unique data structure and learning strategy~\cite{wu2024visionlanguage, xu2024lors}. We first formalize the problem of Multimodal Dataset Distillation (MDD).

\textbf{Problem Formulation.} Given a large-scale image-text dataset $\gD = \{(\vx_i, \bm{\tau}_i), \vy_i\}_{i=1}^{|\gD|}$, 
where $\vx_i \in \sR^{d_{\text{img}}}$ and $\bm{\tau}_i \in \sR^{d_{\text{text}}}$ denote the $i$-th image and its paired caption representation\footnote{Given the discrete nature of text, all subsequent analysis is conducted in the representation space, while images remain processed in the pixel space. Here, $d_{\text{img}}=W\times H\times 3$ and $d_{\text{text}} = 768$ (for BERT~\cite{devlin2018bert}).}, and each pair is independently sampled from a natural data distribution $\mathcal{P}$.
Each $\vy_i \in \{0,1\}^{|\gD|}$ is a one-hot vector indicating the correspondence between $\vx_i$ and the caption set $\{\bm{\tau}_j\}_{j=1}^{|\gD|}$, with the $i$-th entry activated. Similar to DD, MDD also aims to minimize the loss on original dataset using the model trained on its distilled synthetic counterpart $\gS=\{(\tilde{\vx}_i, \tilde{\bm{\tau}}_i), \tilde{\vy}_i \}_{i=1}^{|\gS|}$:
\begin{align}
	\gS^* = \argmin_{\gS} \mathop{\mathbb{E}}_{(\vx,\bm{\tau}) \sim \mathcal{P}} [\mathcal{L}(f_{\bm{\theta}_{\gS}}(\vx,\bm{\tau}), \vy)]\quad \text{s.t.} \quad \bm{\theta}_{\gS} = \argmin_{\bm{\theta}} \mathop{\mathbb{E}}_{(\tilde{\vx},\tilde{\bm{\tau}}) \sim \gS} [\mathcal{L}(f_{\bm{\theta}}(\tilde{\vx},\tilde{\bm{\tau}}), \tilde{\vy})],
	\label{eq:DDobjective}
\end{align} 
where $|\mathcal{S}| \ll |\mathcal{D}|$, and $\mathcal{L}$ denotes the contrastive learning loss. The model $f_{\bm{\theta}}(\cdot)$ represents a CLIP-style network parameterized by $\bm{\theta}$. Each distilled sample consists of a synthetic image-text pair $(\tilde{\vx}_i, \tilde{\bm{\tau}}_i)$, where $\tilde{\vx}_i \in \mathbb{R}^{d_\text{img}}$ and $\tilde{\bm{\tau}}_i \in \mathbb{R}^{d_\text{text}}$, accompanied by a learned soft label $\tilde{\vy}_i$. 
\begin{figure}
    \centering
    \includegraphics[width=1\linewidth]{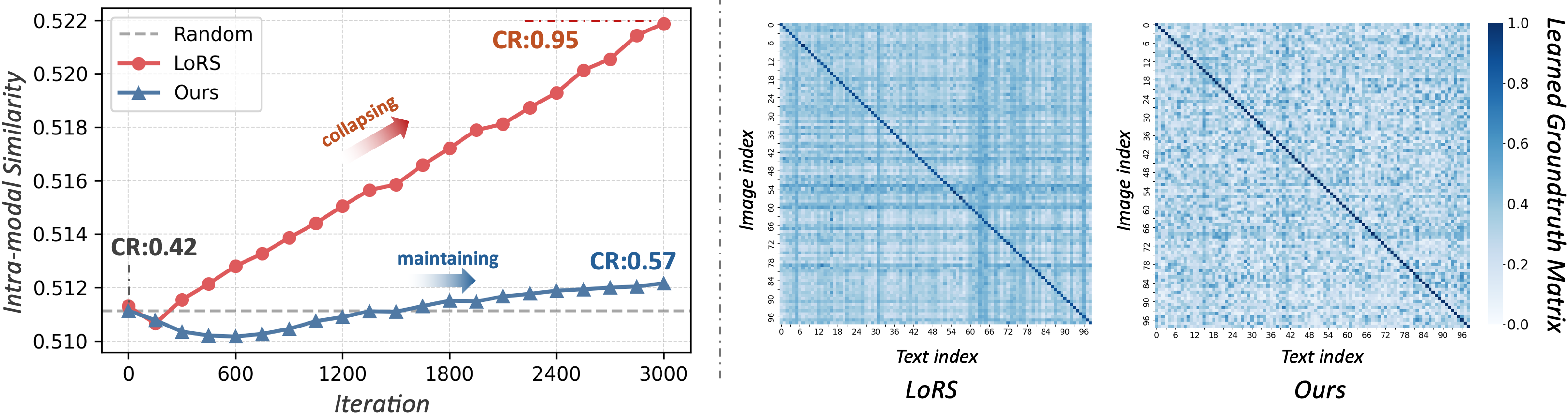}
\vspace{-1.8em}
    \caption{\textbf{Left}: Increasing intra-modal similarity as distillation progresses. We run optimization for 3000 iterations and track the intra-modal cosine similarity, which increases from 0.512 to 0.522 (red curve). Though small in magnitude, this rise leads to a more than twofold increase in concentration ratio (\texttt{CR})\protect\footnotemark{} due to the high dimensionality of the embedding space. \textbf{Right}: Modality collapse undermines the effectiveness of learned soft cross-modal correspondence. The non-matching image-text pairs exhibit nearly uniform similarity scores, forming horizontal and vertical stripes. } 
    \label{fig:problems}
\vspace{-1em}
\end{figure}
\footnotetext{\texttt{CR} measures how tightly the features are clustered, based on how much of the hypersphere is covered at the given cosine similarity. (Refer to \autoref{app:cr} for more calculation details).}

\textbf{MDD vs. Vanilla DD.} According to the \autoref{eq:DDobjective}, the generalization from vanilla DD to MDD involves two key modifications:
1) introducing soft ground-truth vectors $\tilde{\vy}_i$, and
2) optimizing under a contrastive learning loss $\mathcal{L}$ for image-text alignment.
While learning soft labels is common in vanilla DD~\cite{tesla}, optimizing $\tilde{\vy}_i$ in MDD is more challenging, as both image and text representations are updated simultaneously. Besides, in practice, the contrastive loss $\mathcal{L}$ is typically instantiated as InfoNCE~\cite{InfoNCE}, extended InfoNCE (eNCE), or weighted BCE (wBCE)~\cite{xu2024lors}, all aiming to strengthen positive alignments while penalizing mismatched pairs. However, these extensions only make the multimodal adaptation feasible, overlooking the essence of dataset distillation: effective information condensation.
More specifically, they prioritize cross-modal alignment, while failing to preserve intra-modal diversity and discriminability under severe data compression.
\section{Methodology}
In this section, we introduce \textbf{RepBlend}, a novel approach for MDD. We begin by identifying the phenomenon of \textbf{\textit{Modality Collapse}}, which emerges when vanilla DD methods are naively applied to multimodal settings. Through theoretical and empirical analysis, we uncover its underlying causes. To address this issue, we propose {Representation Blending} to enhance intra-modal diversity. In addition, we introduce {Symmetric Projection Trajectory Matching}, which balances the distillation process across modalities and further strengthens cross-modal alignment. The overall pipeline of RepBlend is outlined in \autoref{alg:syn}.
\subsection{Modality Collapse}
LoRS~\cite{xu2024lors} is a representative MDD method built upon \autoref{eq:DDobjective}, where $\mathcal{L}$ is defined as:
\begin{equation}\label{eq:training_loss}
\begin{aligned}
\mathcal{L}_{\mathrm{wBCE}}^\gB = \sum_{i,j}^{|\gB|} w_{ij} \cdot \ell\left(\tilde{\vy}_{ij}, \sigma\left(\hat{{\vy}}_{ij} / \gamma\right)\right), \quad w_{ij} = \frac{\mathbb{I}{[\tilde{\vy}_{ij} > \beta]}}{|\{(i,j): \tilde{\vy}_{ij} > \beta\}|} + \frac{\mathbb{I}{[\tilde{\vy}_{ij} \leq \beta]}}{|\{(i,j): \tilde{\vy}_{ij} \leq \beta\}|}.
\end{aligned}
\end{equation}
Here, $\gB \subset \gS$ denotes a sampled batch. $\hat{\vy}_{ij}$ represents the cosine similarity between the normalized image and text embeddings, where $\tilde{\vx}_i^\prime = \operatorname{Normalize}(f^{\text{imgE}}(\tilde{\vx}_i))$\footnote{In LoRS~\cite{xu2024lors}, no image projection head is used.} and $\tilde{\bm{\tau}}_j^\prime = \operatorname{Normalize}(f^{\text{textP}}(\tilde{\bm{\tau}}_j))$, with $f^{\text{imgE}}(\cdot)$ and $f^{\text{textP}}(\cdot)$ denoting the image encoder and text projection head, respectively. The threshold $\beta$ is used to determine positive and negative pairs, $\sigma(\cdot)$ denotes the sigmoid function, and $\gamma$ is the temperature. $\ell(\cdot, \cdot)$ refers to the binary cross-entropy loss.
While this supervision primarily aims to mine cross-modal relationships, it inadvertently reinforces intra-modal similarities, ultimately leading to \textbf{\textit{Modality Collapse}}, as shown in \autoref{fig:modal_collapse}, where instances within each modality excessively concentrate. Without loss of generality, the following analysis focuses on the image modality.

\textbf{\textit{Proposition: Cross-modal supervision reinforces intra-modal similarity.}} During dataset distillation, if $\{\tilde{\vx}_n, \tilde{\bm{\tau}}_n\}$ and $\{\tilde{\vx}_m, \tilde{\bm{\tau}}_m\}$ exhibit some non-negligible similarity, i.e., $\tilde{\vy}_{nm} \approx \tilde{\vy}_{mn} > \beta$, then the direction of their subsequent updates $\frac{\partial \mathcal{L}}{\partial \tilde{\vx}_n^\prime}\frac{\partial \mathcal{L}}{\partial \tilde{\vx}_m^\prime}$ is determined by
\begin{equation}
\begin{aligned}
 \frac{w_{nm} w_{mn}}{\gamma^2}[\sigma(\hat{\vy}_{nm})/t-\tilde{\vy}_{nm}][\sigma(\hat{\vy}_{mn})/t-\tilde{\vy}_{mn}]{\tilde{\bm{\tau}}}_m^{\prime\top}{\tilde{\bm{\tau}}}_n^{\prime},
\end{aligned}\label{eq:modality_collapse}
\end{equation}
\begin{wrapfigure}{r}{0.5\textwidth}
    \centering
    \includegraphics[width=0.9\linewidth]{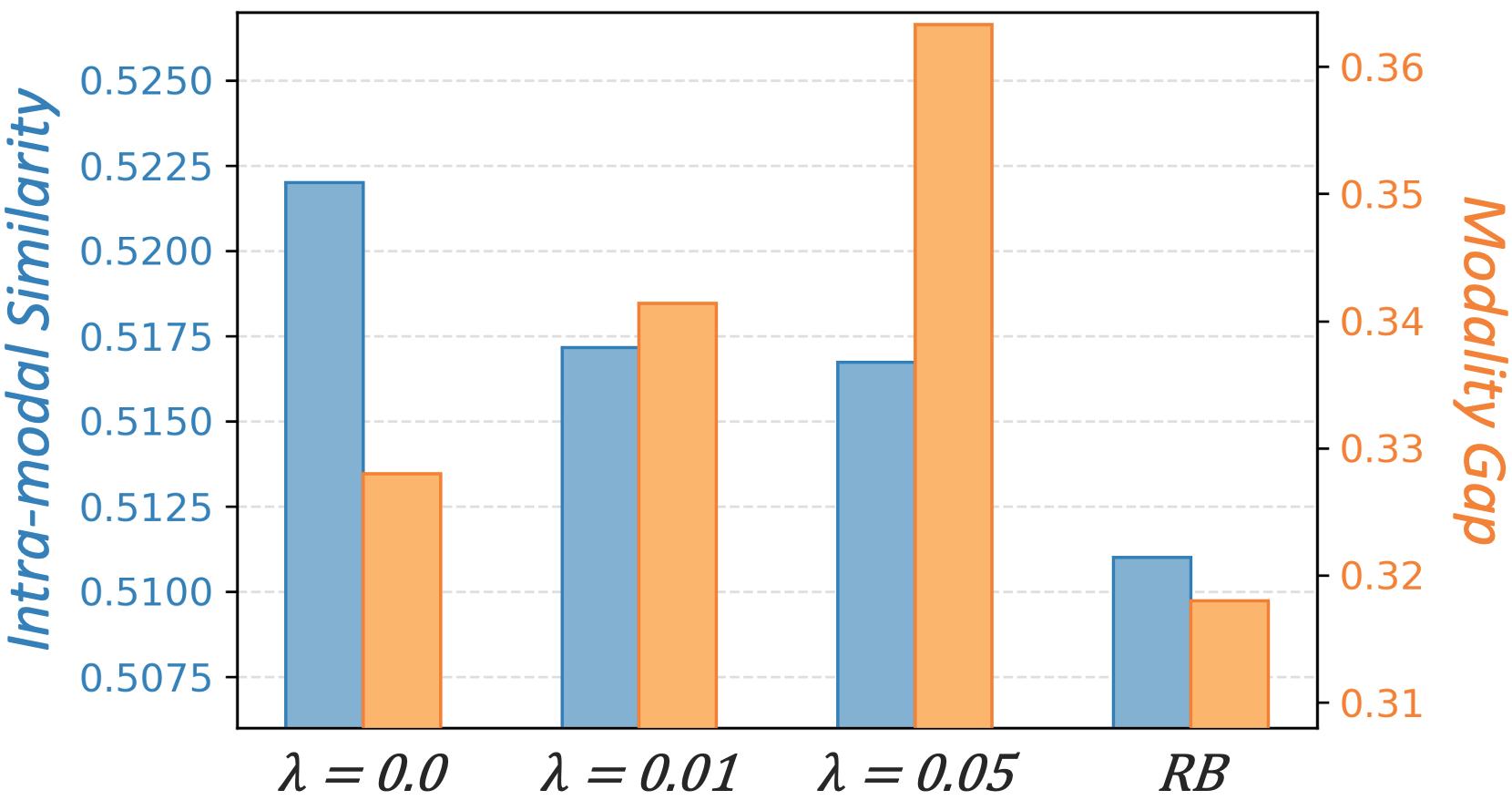}
    \caption{\textcolor{black}{As the noise level $\lambda$ increases, intra-modal similarity (blue bars) shows a slight decline, while the modality gap (yellow bars) rises markedly. In contrast, our representation blending (RB) leverages in-distribution samples to simultaneously reduce intra-modal similarity and inter-modal gap, effectively mitigating modality collapse during distillation.}}
    \label{fig:add_noise}
\vspace{-0.5em}
\end{wrapfigure}
which indicates that the optimization is guided by positive pairs ${\tilde{\bm{\tau}}}_m^{\prime\top}{\tilde{\bm{\tau}}}_n^{\prime}$, promoting concentration in similar directions.
A detailed derivation is provided in \autoref{app:modality_gap}.
When distilling a large dataset into a compact one, the optimization process tends to be dominated by a few salient features~\cite{deng2024iid, dwa2024neurips, datm, shen2024delt}. Once this convergence trend emerges, cross-modal supervision further reinforces it: modality-specific diversity is implicitly suppressed, and intra-modal representations are increasingly aligned toward a limited set of dominant directions. As illustrated in \autoref{fig:problems} (left), \textit{the intra-modal similarity consistently increases throughout the distillation process}.

In addition to the aggravated intra-modal similarity, modality collapse also exacerbates the cross-modal representation gap, as features from each modality become increasingly centralized within compact regions of the shared embedding space. Consequently, the similarities between non-matching image-text pairs converge toward a uniform distribution. Such behavior undermines the utility of soft label distributions, which are designed to encode fine-grained relational information beyond the binary supervision provided by one-hot labels. As illustrated in \autoref{fig:problems} (right), \textit{non-diagonal similarity values exhibit a near-uniform pattern}, where image embeddings produce nearly constant similarity scores across all non-matching text embeddings (manifesting as horizontal stripes), and vice versa for text samples (vertical stripes).  

\subsection{Mitigating Modality Collapse via Representation Blending}
As analyzed in \autoref{eq:modality_collapse}, modality collapse arises from overly strong cross-modal supervision, which implicitly encourages intra-modal concentration and undermines representational diversity. To alleviate this constraint, one potential approach is to inject directional signals that deviate from ${\tilde{\bm{\tau}}}_m^{\prime}$ and ${\tilde{\bm{\tau}}}_n^{\prime}$. 
To empirically validate this hypothesis and explore a viable remedy, we conduct a controlled perturbation experiment on Flickr-30K~\cite{Flickr}. In particular, we adopt two key metrics following \cite{liang2022mind}: the intra-modal similarity ($\texttt{Sim}$) and the modality gap ($\texttt{Gap}$), defined as,
\begin{equation}
\texttt{Sim} = \frac{1}{|\gS|(|\gS| - 1)} \sum_{i \ne j}^{|\gS|} \tilde{\vx}_i^{\prime\top} \tilde{\vx}_j^\prime, \quad
\texttt{Gap} = \frac{1}{|\gS|} \| \sum_{i=1}^{|\gS|} \tilde{\vx}_i^\prime - \sum_{j=1}^{|\gS|} \tilde{\bm{\tau}}_j^\prime \|_2.
\end{equation}
We inject Gaussian noise into the text representations, \\
\resizebox{0.99\linewidth}{!}{$
\begin{aligned}
\tilde{\bm{\tau}}_m^{\prime\text{+noise}} &= \operatorname{Normalize}\left(f^\text{textP}((1-\lambda )\tilde{\bm{\tau}}_m + \lambda \vec{\Delta}_m)\right), \quad
\tilde{\bm{\tau}}_n^{\prime\text{+noise}} &= \operatorname{Normalize}\left(f^\text{textP}((1-\lambda )\tilde{\bm{\tau}}_n + \lambda \vec{\Delta}_n)\right),
\end{aligned}
$} \\ 
where $\vec{\Delta}_m$ and $\vec{\Delta}_n$ are independently sampled random noise from $\mathcal{N}(0,1)$, and $\lambda$ controls the noise level.
We evaluate $\texttt{Sim}$ and $\texttt{Gap}$ under varying levels of $\lambda$. As shown in \autoref{fig:add_noise}, a slight increase in noise reduces intra-modal similarity (blue bars), indicating enhanced modality-specific diversity. These results support our hypothesis that perturbing in the representation space can effectively counteract modality concentration. 

However, as noise level continues to grow, the injected perturbation begins to introduce semantically meaningless signals, which hinders cross-modal alignment. This is evidenced by the growing modality gap (yellow bars), accompanied by a performance drop of 1.9\% in IR@1 and 2.1\% in TR@1 at $\lambda = 0.01$ under 100 distilled pairs on Flickr-30K dataset.
To mitigate this issue, we propose replacing the random perturbation with a structure-preserving variant using in-distribution samples. Specifically, we blends representations from different synthetic instances:
\begin{equation}
\resizebox{0.92\linewidth}{!}{$
\tilde{\bm{\tau}}_m^{\prime\text{blend}} = \operatorname{Normalize}\left(f^\text{textP}((1-\lambda )\tilde{\bm{\tau}}_m + \lambda \tilde{\bm{\tau}}_i)\right), \quad
\tilde{\bm{\tau}}_n^{\prime\text{blend}} = \operatorname{Normalize}\left(f^\text{textP}((1-\lambda )\tilde{\bm{\tau}}_n + \lambda \tilde{\bm{\tau}}_j)\right),
$}
\end{equation}
where $1 \leq i, j \leq |\gS|$. This operation resembles the idea of $\operatorname{MixUp}$, but is applied in the representation space. As shown in the last group of \autoref{fig:add_noise}, we can maintain a low level of intra-modal similarity and small modality gap. Note that although here we illustrate the formulation on text, the same operation is also applied to image side in practice.
\subsection{Enhancing Cross-modal Alignment via Symmetric Projection Trajectory Matching}
In prior MDD practices, methods such as MTT-VL~\cite{wu2024visionlanguage} and LoRS~\cite{xu2024lors} follow a de facto protocol wherein the text encoder is frozen and the image projection layer is omitted. The image encoder and the text projection head are trained to generate expert trajectories for distillation. In this setup, the image encoder is initialized with pretrained weights from ImageNet-1K~\cite{imagenet}, while the text projection head is trained from scratch. This design is motivated by two key considerations: 1) the prohibitive computational and memory cost of optimizing and storing expert trajectories for large-scale text encoders such as BERT~\cite{devlin2018bert}; and 2) the fact that text distillation operates in the representation space, where supervision is applied only through the projection head, thus, matching at the encoder level cannot propagate supervision to the representation space. LoRS~\cite{xu2024lors} minimize the objective in \autoref{eq:DDobjective} through trajectory matching, which is formulated as follows:
\begin{equation*}
\resizebox{0.92\linewidth}{!}{$
\tilde{\vx}^{*}, \tilde{\bm{\tau}}^{*}, \tilde{\vy}^{*} =\underset{\tilde{\vx}, \tilde{\bm{\tau}}, \tilde{\vy}}{\argmin} \left(\left\|{\bm{\theta}}^{t+T}_{\gS_{\text{imgE}}}-\bm{\theta}^{t+M}_{\gD_{\text{imgE}}}\right\|_2^2 +\left\|{\bm{\theta}}^{t+T}_{\gS_{\text{textP}}}-\bm{\theta}^{t+M}_{\gD_{\text{textP}}}\right\|_2^2 \right) \big/ \left(\left\|\bm{\theta}^t_{{\gD_{\text{imgE}}}}-\bm{\theta}^{t+M}_{\gD_{\text{imgE}}}\right\|_2^2 + \left\|\bm{\theta}^t_{{\gD_{\text{textP}}}}-\bm{\theta}^{t+M}_{\gD_{\text{textP}}}\right\|_2^2\right),
$}
\end{equation*}
where \( {\bm{\theta}}^{t+T}_{\gS{\text{imgE}}} \) and \( {\bm{\theta}}^{t+T}_{\gS{\text{textP}}} \) denote the $T$-step finetuned weights of the image encoder and text projection head using $\gS$, initialized from \( \bm{\theta}^t_{{\gD_{\text{imgE}}}} \) and \( \bm{\theta}^t_{{\gD_{\text{textP}}}} \), respectively. The objective is to align the $T$-step synthetic trajectory with the $M$-step real trajectory by minimizing the $\ell_2$ distance between their terminal weights, given the same initialization. 

\begin{figure}
     \centering
     \includegraphics[width=1\linewidth]{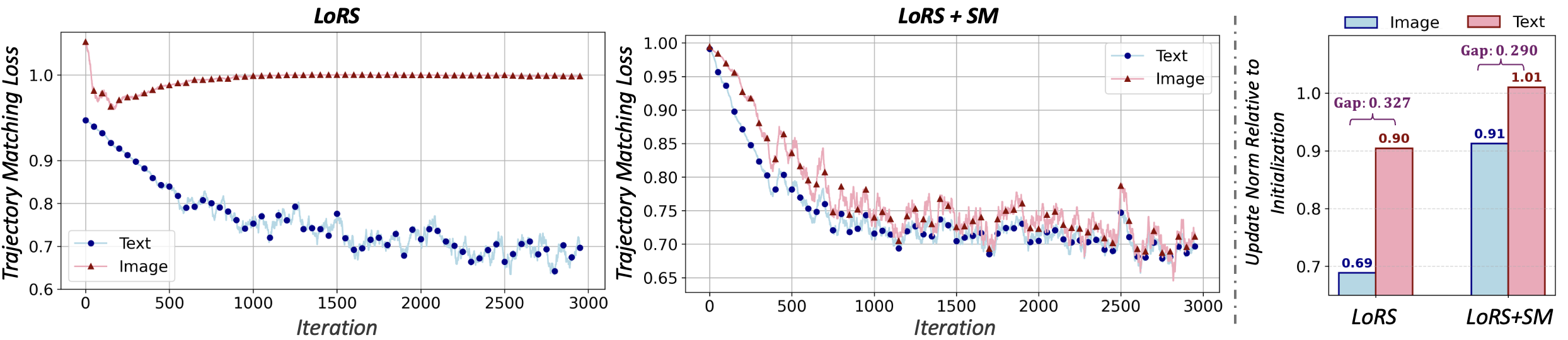}
\vspace{-1.4em}
     \caption{Current MDD methods adopt asymmetric distillation.
\textbf{Left}: The loss on the image side shows much smaller variation than that of the text side, fluctuating mildly around 1.0 without notable reduction.
\textbf{Right}: The update norm relative to initialization is significantly lower for the image modality in LoRS (0.69) compared to the text modality (0.90), suggesting insufficient representation transfer. The update norm is computed in the shared representation space for both modalities.
After incorporating symmetric matching (SM), both image and text modalities exhibit more balanced and synchronized update dynamics, leading to more effective cross-modal alignment (reduced \texttt{Gap}). }
     \label{fig:loss_change}
\vspace{-1em}
\end{figure}

However, the aforementioned trajectory matching is asymmetric. As shown in \autoref{fig:loss_change} (left), the trajectory matching losses of the image and text modalities exhibit divergent trends: the text-side loss decreases steadily, whereas the image-side loss quickly plateaus and remains relatively high. This is primarily because the image encoder contains significantly more parameters than the text projection head, thus, even small per-parameter errors can accumulate into a large overall mismatch. This imbalance is further evidenced in \autoref{fig:loss_change} (right), the norm of updates relative to initialization for the image modality is significantly smaller than that of the text, indicating insufficient distillation on the image side. 
While the representation blending introduced in the previous section helps narrow the modality gap, its effect is still constrained by the inherently asymmetric distillation.
To address this imbalance and further enhance cross-modal alignment, we propose a symmetric distillation strategy by matching trajectories of projection head for both modalities:
\begin{equation}\label{eq:matching_loss}
\resizebox{0.93\linewidth}{!}{$
\tilde{\vx}^{*}, \tilde{\bm{\tau}}^{*}, \tilde{\vy}^{*} =\underset{\tilde{\vx}, \tilde{\bm{\tau}}, \tilde{\vy}}{\argmin} \left(\left\|{\bm{\theta}}^{t+T}_{\gS_{\text{imgP}}}-\bm{\theta}^{t+M}_{\gD_{\text{imgP}}}\right\|_2^2 + \left\|{\bm{\theta}}^{t+T}_{\gS_{\text{textP}}}-\bm{\theta}^{t+M}_{\gD_{\text{textP}}}\right\|_2^2\right) \big/ 
 \left(\left\|\bm{\theta}^t_{{\gD_{\text{imgP}}}}-\bm{\theta}^{t+M}_{\gD_{\text{imgP}}}\right\|_2^2 + \left\|\bm{\theta}^t_{{\gD_{\text{textP}}}}-\bm{\theta}^{t+M}_{\gD_{\text{textP}}}\right\|_2^2\right).
$}
\end{equation}
Here, the image encoder is initialized with ImageNet-1K pretrained weights and kept frozen. While the added image projection head incurs slight computational overhead, it enables projection-based matching that significantly enhances the overall efficiency of the distillation process (as discussed in \textcolor{mylinkcolor}{Section} \ref{sec:exp_compu}).
As shown in \autoref{fig:loss_change}, symmetric projection matching leads to a more consistent decrease in loss for both image and text branches. Moreover, the increased magnitude of updates suggests stronger supervision signals across modalities, resulting in a more balanced and effective distillation process. With symmetric distillation, the modality gap is further narrowed from 0.318 (in \autoref{fig:add_noise}) to 0.290, indicating enhanced cross-modal alignment.
\vspace{-1em}
\begin{algorithm}
\caption{Blending Representations to Mitigate Modality Collapse in MDD}
\label{alg:syn}
\begin{algorithmic}[1]
\Require Original large dataset $\gD$; CLIP-style network $\{f^{\text{imgE}}$, $f^{\text{textE}}$, $f^{\text{imgP}}$, $f^{\text{textP}}\}$; real trajectories set $\bm{\Theta}_{\gD_\text{imgP}}$ and $\bm{\Theta}_{\gD_\text{textP}}$, real trajectory matching length $M$, synthetic trajectory matching length $T$; total optimization iteration number $Iter$
\State Initialize $\gS$ with $|\gS|$ randomly sampled image-text pairs and one-hot groundtruth labels
\State Load pretrained weights into encoders (frozen); randomly initialize projection heads
\For{$it=1$ to $Iter$ } 
\State Sample $\bm{\theta}_{\gD_\text{imgP}}^{t}$, $\bm{\theta}_{\gD_\text{textP}}^{t}$ and $\bm{\theta}_{\gD_\text{imgP}}^{t+M}$, $\bm{\theta}_{\gD_\text{textP}}^{t+M}$ from $\bm{\Theta}_{\gD_\text{imgP}}$ and $\bm{\Theta}_{\gD_\text{textP}}$
\State Initialize $\bm{\theta}_{\gS_\text{imgP}}^{t}$ and $\bm{\theta}_{\gS_\text{textP}}^{t}$ using $\bm{\theta}_{\gD_\text{imgP}}^{t}$ and $\bm{\theta}_{\gD_\text{textP}}^{t}$
\For{$i=1$ to $T$}
\For{mini-batch $\gB=\{(\tilde{\vx}_b, \tilde{\bm{\tau}}_b), \tilde{\vy}_b \}_{b=1}^{|\gB|} \in \gS$}
\State Calculate image representaion $\{f^\text{imgE}(\tilde{\vx}_b)\}$
\State $\triangleright$ \texttt{Blending in representation space}
\vspace{3pt}
\State $\{f^\text{imgE}(\tilde{\vx}_b), \tilde{\bm{\tau}}_b\}$ = $\operatorname{RepBlend}(\{f^\text{imgE}(\tilde{\vx}_b), \tilde{\bm{\tau}}_b\})$ 
\State Compute loss $\mathcal{L}_{\mathrm{wBCE}}^{\gB}$ using \autoref{eq:training_loss} 
\State Update projection head weights $\bm{\theta}_{\gS_\text{imgP}}^{t+i}$ and $\bm{\theta}_{\gS_\text{textP}}^{t+i}$
\EndFor
\State $\triangleright$ \texttt{Symmetric projection trajectory matching}
\State Optimize $\gS=\{(\tilde{\vx}_j, \tilde{\bm{\tau}}_j), \tilde{\vy}_j \}_{j=1}^{|\gS|}$ according to \autoref{eq:matching_loss}
\EndFor
\EndFor
\Ensure Synthetic dataset $\gS$
\end{algorithmic}
\end{algorithm}
\section{Experiments}
In this section, we conduct extensive experiments on multiple benchmark datasets to demonstrate the effectiveness of the proposed RepBlend framework. We first present the experimental setup, including the datasets, baseline methods, and implementation details. The main results are summarized in \autoref{tab:result-main-flickr} and \autoref{tab:result-main-coco}. In addition, we also provide detailed ablation studies to evaluate the individual contribution of each component. All experiments are conducted using two NVIDIA RTX 3090 GPUs and one NVIDIA H100 GPU.
\vspace{-1em}
\subsection{Experimental Setup}
\noindent{\textbf{Datasets and Networks.}}
We evaluate our method on two widely-used image captioning datasets: Flickr-30K~\cite{Flickr} and MS-COCO~\cite{COCO}, which contain approximately 31k and 123k images respectively, with each image paired with five human-annotated captions. For the image encoder, we experiment with NFNet~\cite{brock2021high}, RegNet~\cite{regnet}, ResNet-50~\cite{resnet}, and ViT~\cite{vit}. For the text encoder, we consider both BERT~\cite{devlin2018bert} and DistilBERT~\cite{sanh2019distilbert}.
\textcolor{black}{To further demonstrate the generalizability of our approach across modalities, we extend our evaluation to the AudioCaps~\cite{audiocaps} audio-text benchmark, utilizing 
EfficientAT
~\cite{schmid2023efficient} as the audio encoder. }
Model performance is primarily evaluated using Recall at K (R@K) in cross-modal retrieval tasks. Given a query from one modality, we retrieve the top-$K$ most similar samples from the other modality and measure the retrieval accuracy. We denote text-to-image retrieval as IR@K, and image-to-text retrieval as TR@K.

\noindent{\textbf{Baselines.}} The comparison encompasses a range of state-of-the-art approaches, including coreset selection methods such as Random sampling, Herding~\cite{c-herd}, K-Center~\cite{c-kcenter}, and Forgetting~\cite{c-forget}, as well as recent advances in dataset distillation tailored for vision-language models, including MTT-VL~\cite{wu2024visionlanguage}, TESLA-VL~\cite{xu2024lors}, and LoRS~\cite{xu2024lors}. A detailed description of these methods can be found in the \autoref{app:comparison_methods}.

\noindent{\textbf{Implementation Details.}}
We construct a CLIP-style architecture using the aforementioned image and text encoders. The image encoder is initialized with ImageNet-pretrained weights~\cite{imagenet}, while the text encoder is initialized with the official pretrained weights provided by the corresponding language model. After feature extraction, the outputs from both branches are passed through separate linear projection layers to obtain the final embeddings.
During buffer generation, distillation, and evaluation training, the encoders are frozen and only the projection layers are optimized. We collect 20 expert trajectories, each consisting of 10 training epochs. The hyperparameter settings follow those used in LoRS~\cite{xu2024lors} and can be found in \autoref{tab:Hyperparameter_buffer} and \autoref{tab:Hyperparameter_distillation} in \autoref{app:hyperparameter_settings}.
\begin{table*}[ht]
\vspace{-0.5em}
\caption{Results on Flickr-30k~\cite{Flickr}. Both distillation and validation are performed using NFNet+BERT. The model trained on full dataset performs: IR@1=23.16, IR@5=53.98, IR@10=66.62; TR@1=33.8, TR@5=65.7, TR@10=76.9. For fairness, both LoRS~\cite{xu2024lors} and ours synthesize one fewer pair under each distillation budget (e.g., 99 pairs for a budget of 100)\protect\footnotemark{}.}
\vspace{-0.75em}
\label{tab:result-main-flickr}
\begin{center}
\begin{small}
\resizebox{0.99\textwidth}{!}{%
\begin{tabular}{lcl|c@{\hskip 4.2pt}c@{\hskip 4.2pt}c@{\hskip 4.2pt}c|cccc}
    \toprule
    \multirow{2}{*}{Pairs} & \multirow{2}{*}{Ratio} & \multirow{2}{*}{Metric} & \multicolumn{4}{c|}{Coreset Selection} & \multicolumn{4}{c}{Dataset Distillation} \\
    \cmidrule(lr){4-11}
      &  &  & Rand & Herd~\cite{c-herd} & K-Cent~\cite{c-kcenter} & Forget~\cite{c-forget} & MTT-VL~\cite{wu2024visionlanguage} & TESLA-VL~\cite{xu2024lors} & LoRS~\cite{xu2024lors}& Ours \\
    \midrule
    \multirow{6}{*}{\shortstack{100}} & \multirow{6}{*}{0.3\%}
    & IR@1  & 1.0  & 0.7 & 0.7 & 0.7 &  \metric{4.7}{0.2} &  \metric{0.5}{0.2}  &  \metric{8.3}{0.2}&\textbf{\metricbl{11.5}{0.4}} \\
    & & IR@5  & 4.0  & 2.8 & 3.1 & 2.4 & \metric{15.7}{0.5} &  \metric{2.3}{0.2}  &  \metric{24.1}{0.2}&\textbf{\metricbl{32.0}{0.7}} \\
    & & IR@10 & 6.5  & 5.3 & 6.1 & 5.6 & \metric{24.6}{1.0} &  \metric{4.7}{0.4}  &  \metric{35.1}{0.3} &\textbf{\metricbl{44.5}{0.6}}\\
    & & TR@1  & 1.3  & 1.1 & 0.6 & 1.2 &  \metric{9.9}{0.3} &  \metric{5.5}{0.5}  &  \metric{11.8}{0.2} &\textbf{\metricbl{16.2}{0.8}}\\
    & & TR@5  & 5.9  & 4.7 & 5.0 & 4.2 & \metric{28.3}{0.5} & \metric{19.5}{0.9}  &  \metric{35.8}{0.6}&\textbf{\metricbl{41.7}{0.9}} \\
    & & TR@10 & 10.1 & 7.9 & 7.6 & 9.7 & \metric{39.1}{0.7} & \metric{28.9}{1.0}  &  \metric{49.2}{0.5}&\textbf{\metricbl{55.5}{0.4}} \\
    \midrule
    \multirow{6}{*}{\shortstack{200}} & \multirow{6}{*}{0.7\%}
    & IR@1  & 1.1  & 1.5  & 1.5  & 1.2  &  \metric{4.6}{0.9} &  \metric{0.2}{0.1}  &  \metric{8.6}{0.3} &\textbf{\metricbl{12.7}{0.8}}\\
    & & IR@5  & 4.8  & 5.5  & 5.4  & 3.1  & \metric{16.0}{1.6} &  \metric{1.3}{0.2}  &  \metric{25.3}{0.2} &\textbf{\metricbl{34.7}{0.6}}\\
    & & IR@10 & 9.2  & 9.3  & 9.9  & 8.4  & \metric{25.5}{2.6} &  \metric{2.5}{0.2}  &  \metric{36.6}{0.3} &\textbf{\metricbl{47.6}{0.5}}\\
    & & TR@1  & 2.1  & 2.3  & 2.2  & 1.5  & \metric{10.2}{0.8} &  \metric{2.8}{0.5}  &  \metric{14.5}{0.5} &\textbf{\metricbl{18.6}{0.7}}\\
    & & TR@5  & 8.7  & 8.4  & 8.2  & 8.4  & \metric{28.7}{1.0} & \metric{10.4}{1.5}  &  \metric{38.7}{0.5} &\textbf{\metricbl{46.0}{0.8}}\\
    & & TR@10 & 13.2 & 14.4 & 13.5 & 10.2 & \metric{41.9}{1.9} & \metric{17.4}{1.6}  &  \metric{53.4}{0.5} &\textbf{\metricbl{60.0}{0.6}}\\
    \midrule
    \multirow{6}{*}{\shortstack{500}} & \multirow{6}{*}{1.7\%}
      & IR@1  & 2.4  &3.0  & 3.5  & 1.8  & \metric{6.6}{0.3} &  \metric{1.1}{0.2}  & \metric{10.0}{0.2} &\textbf{\metricbl{17.0}{0.6}}\\ 
    & & IR@5  & 10.5 &10.0 & 10.4 & 9.0  & \metric{20.2}{1.2} &  \metric{7.3}{0.4}  & \metric{28.9}{0.7} &\textbf{\metricbl{42.5}{0.5}}\\ 
    & & IR@10 & 17.4 &17.0 & 17.3 & 15.9 & \metric{30.0}{2.1} & \metric{12.6}{0.5}  & \metric{41.6}{0.6} &\textbf{\metricbl{55.9}{0.6}}\\ 
    & & TR@1  & 5.2  &5.1  & 4.9  & 3.6  & \metric{13.3}{0.6} &  \metric{5.1}{0.2}  & \metric{15.5}{0.7} &\textbf{\metricbl{22.5}{0.4}}\\ 
    & & TR@5  & 18.3 &16.4 & 16.4 & 12.3 & \metric{32.8}{1.8} & \metric{15.3}{0.5}  & \metric{39.8}{0.4} &\textbf{\metricbl{53.2}{0.3}}\\ 
    & & TR@10 & 25.7 &24.3 & 23.3 & 19.3 & \metric{46.8}{0.8} & \metric{23.8}{0.3}  & \metric{53.7}{0.3} &\textbf{\metricbl{66.7}{0.3}}\\
    \bottomrule
    \end{tabular}%
}
\end{small}
\end{center}
\vspace{-8pt}
\end{table*} 
\footnotetext{To offset the additional memory overhead introduced by soft labels.}
\subsection{Main Results}
The results on Flickr-30K~\cite{Flickr} and MS-COCO~\cite{COCO} are presented in \autoref{tab:result-main-flickr} and \autoref{tab:result-main-coco}, respectively. Our method consistently outperforms all baseline methods, across all distillation budgets and evaluation metrics.
Notably, on Flickr-30k, under the extremely low-data regime of 100 training pairs (0.3\%), our method achieves an IR@1 of 11.5\%, substantially surpassing LoRS (8.3\%) and MTT-VL (4.7\%). Similarly, our TR@10 reaches 55.5\%, a considerable gain over the best baseline LoRS (49.2\%). These trends hold consistently across all pair settings. Under the 500-pair scenario (1.7\%), our method improves the IR@10 from 41.6\% (LoRS) to 55.9\% and TR@10 from 53.7\% to 66.7\%, reflecting a relative gain of over 30\%.
On MS-COCO, a dataset known for higher complexity and variability, our method continues to exhibit superior performance. Under the 100-pair setting (0.8‰), our approach achieves IR@10 = 22.3\% and TR@10 = 28.0\%, substantially outperforming LoRS, which attains 12.2\% and 19.6\%, respectively. At a higher budget of 500 training pairs (4.4‰), our method maintains its advantage, achieving the highest IR@10 (30.6\%) and TR@10 (32.9\%) among all evaluated methods.
The observed improvements are both substantial and consistent, demonstrating the effectiveness of our distillation framework in condensing multimodal datasets. \textcolor{black}{Moreover, our method also demonstrates strong generalizability to other multimodal settings, such as audio-text benchmark. See \autoref{app:generalization_to_audio_text} for details.}
\begin{table*}[ht]
\vspace{-0.75em}
\caption{Results on MS-COCO~\cite{COCO}. Both distillation and validation are performed using NFNet+BERT. The model trained on full dataset performs: IR@1=14.6, IR@5=38.9, IR@10=53.2; TR@1=20.6, TR@5=46.8, TR@10=61.3. For fairness, both LoRS~\cite{xu2024lors} and ours synthesize one fewer pair under each distillation budget (e.g., 99 pairs for a budget of 100).}
\vspace{-0.9em}
\label{tab:result-main-coco}
\begin{center}
\begin{small}
\resizebox{0.99\textwidth}{!}{%
\begin{tabular}{lcl|c@{\hskip 4.2pt}c@{\hskip 4.2pt}c@{\hskip 4.2pt}c|cccc}
    \toprule
    \multirow{2}{*}{Pairs} & \multirow{2}{*}{Ratio} & \multirow{2}{*}{Metric} & \multicolumn{4}{c|}{Coreset Selection} & \multicolumn{4}{c}{Dataset Distillation} \\
    \cmidrule(lr){4-11}
      &  &  & Rand & Herd~\cite{c-herd} & K-Cent~\cite{c-kcenter} & Forget~\cite{c-forget} & MTT-VL~\cite{wu2024visionlanguage} & TESLA-VL~\cite{xu2024lors} & LoRS~\cite{xu2024lors}& Ours \\
    \midrule
    \multirow{6}{*}{\shortstack{100}} & \multirow{6}{*}{0.8\textperthousand}
    & IR@1  & 0.3 & 0.5 & 0.4 & 0.3 &  \metric{1.3}{0.1} & \metric{0.3}{0.2}  & \metric{1.8}{0.1} & \textbf{\metricbl{4.1}{0.3}} \\ 
    & & IR@5  & 1.3 & 1.4 & 1.4 & 1.5 &  \metric{5.4}{0.3} & \metric{1.0}{0.4}  & \metric{7.1}{0.2}&\textbf{\metricbl{13.9}{0.8}} \\ 
    & & IR@10 & 2.7 & 3.5 & 2.5 & 2.5 &  \metric{9.5}{0.5} & \metric{1.8}{0.5}  & \metric{12.2}{0.2}&\textbf{\metricbl{22.3}{0.5}} \\ 
    & & TR@1  & 0.8 & 0.8 & 1.4 & 0.7 &  \metric{2.5}{0.3} & \metric{2.0}{0.2}  & \metric{3.3}{0.2}&\textbf{\metricbl{5.2}{0.5}} \\ 
    & & TR@5  & 3.0 & 2.1 & 3.7 & 2.6 & \metric{10.0}{0.5} & \metric{7.7}{0.5}  & \metric{12.2}{0.3}&\textbf{\metricbl{17.9}{0.9}} \\ 
    & & TR@10 & 5.0 & 4.9 & 5.5 & 4.8 & \metric{15.7}{0.4} & \metric{13.5}{0.3}  & \metric{19.6}{0.3}&\textbf{\metricbl{28.0}{0.3}} \\
    \midrule
    \multirow{6}{*}{\shortstack{200}} & \multirow{6}{*}{1.7\textperthousand}
    & IR@1  & 0.6 & 0.9 & 0.7 & 0.6 &  \metric{1.7}{0.1} &  \metric{0.1}{0.1} &  \metric{2.4}{0.1}&\textbf{\metricbl{6.1}{0.8}} \\
    & & IR@5  & 2.3 & 2.4 & 2.1 & 2.8 &  \metric{6.5}{0.4} &  \metric{0.2}{0.1} &  \metric{9.3}{0.2}&\textbf{\metricbl{19.3}{0.7}} \\
    & & IR@10 & 4.4 & 4.1 & 5.8 & 4.9 & \metric{12.3}{0.8} &  \metric{0.5}{0.1} &  \metric{15.5}{0.2}&\textbf{\metricbl{29.8}{0.5}} \\
    & & TR@1  & 1.0 & 1.0 & 1.2 & 1.1 &  \metric{3.3}{0.2} &  \metric{0.7}{0.2} &  \metric{4.3}{0.1}&\textbf{\metricbl{6.9}{0.6}} \\
    & & TR@5  & 4.0 & 3.6 & 3.8 & 3.5 & \metric{11.9}{0.6} &  \metric{3.1}{0.5} &  \metric{14.2}{0.3}&\textbf{\metricbl{21.8}{0.9}} \\
    & & TR@10 & 7.2 & 7.7 & 7.5 & 7.0 & \metric{19.4}{1.2} &  \metric{5.3}{0.8} &  \metric{22.6}{0.2}&\textbf{\metricbl{32.3}{0.7}} \\
    \midrule
    \multirow{6}{*}{\shortstack{500}} & \multirow{6}{*}{4.4\textperthousand}
    & IR@1  & 1.1 & 1.7 & 1.1  & 0.8 &  \metric{2.5}{0.5}  &  \metric{0.8}{0.2} & \metric{2.8}{0.2}&\textbf{\metricbl{6.2}{0.1}} \\
    & & IR@5  & 5.0 & 5.3 & 6.3  & 5.8 &  \metric{8.9}{0.7}  &  \metric{3.6}{0.6} & \metric{9.9}{0.5}&\textbf{\metricbl{19.9}{0.3}} \\
    & & IR@10 & 8.7 & 9.9 & 10.5 & 8.2 & \metric{15.8}{1.5}  &  \metric{6.7}{0.9} & \metric{16.5}{0.7}&\textbf{\metricbl{30.6}{0.1}} \\
    & & TR@1  & 1.9 & 1.9 & 2.5  & 2.1 &  \metric{5.0}{0.4}  &  \metric{1.7}{0.4} & \metric{5.3}{0.5} &\textbf{\metricbl{7.0}{0.2}}\\
    & & TR@5  & 7.5 & 7.8 & 8.7  & 8.2 & \metric{17.2}{1.3}  &  \metric{5.9}{0.8} & \metric{18.3}{1.5}&\textbf{\metricbl{22.0}{0.3}} \\
    & & TR@10 & 12.5& 13.7& 14.3 & 13.0& \metric{26.0}{1.9}  & \metric{10.2}{1.0} & \metric{27.9}{1.4}&\textbf{\metricbl{32.9}{0.6}} \\
    \bottomrule
    \end{tabular}%
}
\end{small}
\end{center}
\vspace{-2em}
\end{table*}
\begin{figure}
    \centering
    \includegraphics[width=0.99\linewidth]{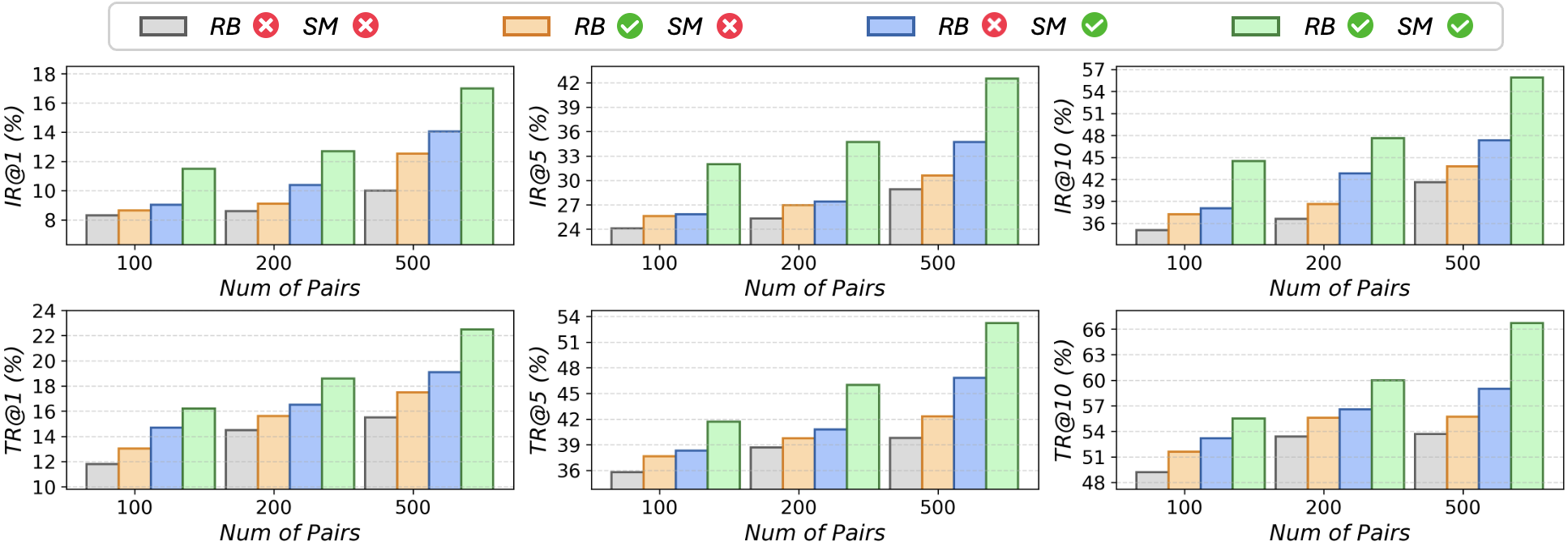}
\vspace{-0.5em}
    \caption{Ablation study of Representation Blending (RB) and Symmetric Projection Trajectory Matching (SM) on Flickr-30K with NFNet+BERT. }
    \label{fig:ablation_study_flickr}
\end{figure}
\textcolor{black}{
\subsection{Ablation Study}
\noindent{\textbf{Representation Blending \&  Symmetric Matching.}}
We conduct an ablation study on the Flickr-30K dataset using NFNet+BERT to evaluate the individual and combined contributions of the proposed components: Representation Blending (RB) and Symmetric Projection Trajectory Matching (SM). As shown in \autoref{fig:ablation_study_flickr}, removing either module leads to consistent performance degradation across all retrieval metrics (IR@1/5/10 and TR@1/5/10) and distillation budgets (100, 200, 500 pairs).
RB contributes by mitigating intra-modal collapse; as illustrated in \autoref{fig:add_noise}, it effectively reduces intra-modal similarity and enhances representational diversity.
SM further balances the learning dynamics across modalities and improves cross-modal alignment, as evidenced in \autoref{fig:loss_change}.
When combined, RB and SM achieve the best overall performance, highlighting their complementary roles in enhancing intra-modal diversity and cross-modal alignment.
}

\noindent{\textbf{Cross-Architecture Generalization.}}
We further validate the generalization capability of RepBlend across diverse architectures. Following the protocol of LoRS~\cite{xu2024lors}, we keep the text encoder fixed and evaluate the dataset distilled with NFNet+BERT using alternative image encoders, including ResNet-50 and RegNet. As shown in \autoref{tab:cross-arch}, RepBlend consistently maintains strong performance across different encoder architectures.
Moreover, we extend the evaluation to a broader set of architecture combinations, such as ResNet-50+BERT, ViT+BERT, RegNet+BERT, and NFNet+DistilBERT, as illustrated in \autoref{fig:multiarch_Flickr} and \autoref{fig:multiarch_coco} in \autoref{app:multi_arch}. Across all architectures, datasets, and distillation budgets, RepBlend consistently outperforms the sota baseline, demonstrating its robustness and architectural adaptability.
\begin{table*}[t]
\caption{Cross-architecture generalization. The distilled data are synthesized using NFNet+BERT and evaluated across different architectures. Evaluations are conducted on Flickr-30K under the 500-pair setting. For fairness, both LoRS~\cite{xu2024lors} and ours synthesize one fewer pair, e.g., 499 pairs.}
\vspace{-0.5em}
\label{tab:cross-arch}
\begin{center}
\begin{small}
\begin{sc}
\resizebox{0.93\linewidth}{!}{%
    \begin{tabular}{ll|cccccc}
    \toprule
      {Evaluate Model} & {Methods} & IR@1 & IR@5 & IR@10& TR@1 & TR@5 & TR@10 \\ 
    \midrule
    \multirow{3}{*}{ResNet+BERT}
   & TESLA-VL~\cite{xu2024lors}  & \metric{3.0}{0.2} & \metric{10.8}{0.5} & \metric{17.0}{0.8} & \metric{6.0}{0.9} & \metric{18.8}{0.7} & \metric{27.7}{1.2}  \\
    & LoRS~\cite{xu2024lors}  & \metric{3.3}{0.2} & \metric{12.7}{0.3} & \metric{20.4}{0.2} & \metric{6.8}{0.2} & \metric{19.6}{1.3} & \metric{31.1}{0.3}  \\
    & Ours & \textbf{\metricbl{4.2}{0.2}}&\textbf{\metricbl{14.1}{0.2}}&\textbf{\metricbl{23.6}{0.6}}&\textbf{\metricbl{8.4}{0.2}} & \textbf{\metricbl{23.1}{0.8}} &\textbf{\metricbl{35.0}{1.3}}\\
    \midrule
     \multirow{3}{*}{RegNet+BERT}
     & TESLA-VL~\cite{xu2024lors}   & \metric{3.2}{0.8} & \metric{11.1}{1.8} & \metric{17.5}{1.3} & \metric{5.8}{0.1} & \metric{18.6}{0.6} & \metric{28.1}{1.0} \\
    & LoRS~\cite{xu2024lors}  & \metric{3.5}{0.1} & \metric{12.6}{0.3} & \metric{21.1}{0.4} & \metric{6.8}{0.3} & \metric{20.8}{0.3} & \metric{30.2}{0.3} \\
    & Ours & \textbf{\metricbl{3.9}{0.2}}&\textbf{\metricbl{13.9}{0.3}}&\textbf{\metricbl{24.0}{0.6}}&\textbf{\metricbl{7.9}{0.3}} & \textbf{\metricbl{24.2}{0.3}} &\textbf{\metricbl{36.2}{1.1}}\\
    \bottomrule
    \end{tabular}%
}
\end{sc}
\end{small}
\end{center}
\vspace{-2em}
\end{table*}
\subsection{Computational Efficiency}\label{sec:exp_compu}
\begin{wraptable}{r}{0.42\textwidth}
\vspace{-4em}
\caption{Study of computational efficiency.}
\centering
\scriptsize
\begin{tabular}{>{\centering\arraybackslash}p{2.4cm}|>{\centering\arraybackslash}p{1.2cm}>{\centering\arraybackslash}p{1.0cm}}
    \toprule
    {Methods} & {LoRS~\cite{xu2024lors}} & {Ours} \\
    \midrule
    (IR@1, TR@1) (\%) & (8.3, 11.8) & $\metricblbold{(11.5, 16.2)}$ \\
    \midrule
    \rowcolor{gray!15}
    \multicolumn{3}{c}{{Buffer}} \\
    \noalign{\vskip -1.0mm}  
    \midrule
    Speed (min/traj) & 70 & $\metricblbold{40}$ \\
    Memory (GB/traj) & 1.63 & $\metricblbold{0.73}$ \\
    \midrule
    \rowcolor{gray!15}
    \multicolumn{3}{c}{{Distillation}} \\
    \noalign{\vskip -1.0mm}  
    \midrule
    Speed (s/iter) & 11.5 & $\metricblbold{1.71}$ \\
    Peak GPU VRAM (GB) & 21.78 & $\metricblbold{10.17}$ \\
    \bottomrule
\end{tabular}
\label{tab:efficiency}
\vspace{-1.5em} 
\end{wraptable}
In the proposed method, the training trajectories of image and text projection layers are used for matching optimization. Although we introduce an additional image projection, it incurs negligible computational overhead. In fact, as shown in \autoref{tab:efficiency}, our method achieves significantly better computational efficiency compared to prior work. Specifically, the time required to construct expert trajectories is reduced from 70 minutes to 40 minutes per trajectory (1.75$\times$ speedup), and the corresponding memory footprint decreases from 1.63 GB to 0.73 GB (2.23$\times$ reduction). During the distillation phase, our method accelerates training iterations from 11.5 seconds to 1.71 seconds per iteration, yielding a 6.7$\times$ speedup. Moreover, it lowers the peak GPU memory usage from 21.78 GB to 10.17 GB (2.14$\times$ reduction). These results demonstrate that our projection-based design not only enables more effective multimodal distillation, but also leads to substantially improved computational efficiency.

\section{Conclusion}
\textcolor{black}{
In this work, we investigate the underexplored challenge of modality collapse in multimodal dataset distillation (MDD), where intra-modal similarity is excessively amplified and inter-modal alignment is degraded. Through theoretical analysis and empirical evidence, we attribute this phenomenon to the inherent over-compression behavior of dataset distillation and its interplay with cross-modal contrastive supervision. To mitigate these issues, we propose RepBlend, a novel MDD framework incorporating two key components: Representation Blending for enhancing intra-modal diversity and Symmetric Projection Trajectory Matching for achieving balanced and effective supervision across modalities. Extensive experiments on Flickr-30K and MS-COCO confirm the superiority of RepBlend in both retrieval performance and distillation efficiency.}

\noindent\textbf{Limitations and Future work.} Despite the promising results of RepBlend, current MDD frameworks, including ours, remain limited to pair-level modeling, which restricts fine-grained alignment between text tokens and visual objects. Additionally, insufficient cross-instance interaction hampers representation expressiveness and limits further gains in compression. In the future, we will explore instance-aware, relation-enhanced strategies to overcome these challenges.

\newpage
\bibliography{neurips_2025}
\bibliographystyle{plain}

\clearpage
\newpage
\appendix
\section{More Related Works}
Dataset distillation (DD), first proposed by Wang et al.~\cite{dd2018}, aims to improve training efficiency by condensing information from large-scale datasets into a small set of synthetic samples. Building on this foundation, recent advancements have introduced a wide range of techniques for effectively and efficiently compressing representative knowledge into compact datasets. Depending on the underlying distillation objective, existing DD methods can be broadly categorized into gradient matching~\cite{dc2021, dsa2021, lee2022dataset, shin2023loss}, trajectory matching~\cite{mtt, tesla, FTD, seqmatch}, and distribution matching~\cite{wang2022cafe, DM, sajedi2023datadam, sun2024diversity, deng2024iid, sre, yin2023dataset}. 
Among these, trajectory matching approaches demonstrate competitive performance without relying on additional label augmentation, making them particularly effective and efficient for practical distillation tasks.

While early efforts have predominantly focused on image data, recent works have extended DD to other domains such as text~\cite{lu2025unidetox,maekawa2025dilm}, video~\cite{ding2025condensing, wang2024dancing}, and graph data~\cite{liu2023graph, zhang2024navigating}. For example, DiLM~\cite{maekawa2025dilm} leverages a generative language model to produce textual synthetic data, enabling model-agnostic distillation with strong generalization. Wang et al.~\cite{wang2024dancing} address the underexplored challenge of temporal compression in videos by disentangling spatial and temporal information. In the graph domain, GDEM~\cite{liu2023graph} aligns the eigenbasis and node features of real and synthetic graphs, achieving efficient and architecture-agnostic distillation without relying on GNN-specific supervision.
These promising achievements naturally motivate exploration into multimodal scenarios. MTT-VL~\cite{wu2024visionlanguage} is the first attempt in this direction, adapting trajectory matching for image-text datasets and demonstrating the feasibility of distilling multimodal information. Building upon this, LoRS~\cite{xu2024lors} further investigates the unique challenge in multimodal dataset distillation (MDD), i.e., high representational variance, and proposes to construct a similarity matrix to mine associations between all matched and mismatched pairs more effectively. Despite these advances, existing methods remain focused on data structures, overlooking the fundamental impact of contrastive objectives in multimodal optimization, which can lead to modality collapse. In this paper, we propose an effective and efficient MDD framework that explicitly addresses this issue.
\section{Derivation of \autoref{eq:modality_collapse}}\label{app:modality_gap}
As defined in \autoref{eq:training_loss}, 
\begin{equation*}
\begin{aligned}
\mathcal{L}_{\mathrm{wBCE}}^\gB = \sum_{i,j}^{|\gB|} w_{ij} \cdot \ell\left(\tilde{\vy}_{ij}, \sigma\left(\hat{{\vy}}_{ij} / \gamma\right)\right), \quad w_{ij} = \frac{\mathbb{I}{[\tilde{\vy}_{ij} > \beta]}}{|\{(i,j): \tilde{\vy}_{ij} > \beta\}|} + \frac{\mathbb{I}{[\tilde{\vy}_{ij} \leq \beta]}}{|\{(i,j): \tilde{\vy}_{ij} \leq \beta\}|},
\end{aligned}
\end{equation*}
where $\sigma(x)$ is the sigmoid function and $\ell(y,p)=-y\log(p)-(1-y)\log(1-p)$ is the binary cross-entropy loss. Thus, we have:
\begin{equation*}
\begin{aligned}
\ell\left(y,\sigma(x)\right) 
& = -y\log\frac1{1+e^{-x}}-(1-y)\log\frac{e^{-x}}{1+e^{-x}} \\
& = y\log(1+e^{-x})+(1-y)x+(1-y)\log(1+e^{-x})=\log(1+e^{-x})+(1-y)x,
\end{aligned}
\end{equation*}
whose derivative with respect to $x$ is: 
\begin{equation*}
\begin{aligned}
& \frac{\partial\ell\left(y,\sigma(x)\right) }{\partial x}=\frac{-e^{-x}}{1+e^{-x}}+(1-y)=\sigma(x)-y.
\end{aligned}
\end{equation*}
Given $\hat{\vy}_{ij}=\tilde{\vx}_i^{\prime\top} \tilde{\bm{\tau}}_j^\prime $, the overall gradient of wBCE is:
\begin{equation*}
\frac{\partial \mathcal{L}}{\partial \tilde{\vx}_n^\prime}
=  \sum_{j=1}^{|\gB|}w_{nj}\frac{\partial} {\partial \tilde{\vx}_n^\prime} \ell\left(\tilde{\vy}_{nj}, \sigma(\tilde{\vx}_n^{\prime\top} \tilde{\bm{\tau}}_j^\prime/\gamma)\right) 
=  \sum_{j=1}^{|\gB|}\frac{w_{nj}}{\gamma}(\sigma(\hat{\vy}_{nj}/\gamma)-\tilde{\vy}_{nj})\tilde{\bm{\tau}}_j^\prime. 
\end{equation*}
Similarly,
\begin{equation*}
\frac{\partial \mathcal{L}}{\partial \tilde{\vx}_m^\prime}
=  \sum_{j=1}^{|\gB|}w_{mj}\frac{\partial} {\partial \tilde{\vx}_m^\prime} \ell\left(\tilde{\vy}_{mj}, \sigma(\tilde{\vx}_m^{\prime\top} \tilde{\bm{\tau}}_j^\prime/\gamma)\right) 
=  \sum_{j=1}^{|\gB|}\frac{w_{mj}}{\gamma}(\sigma(\hat{\vy}_{mj}/\gamma)-\tilde{\vy}_{mj})\tilde{\bm{\tau}}_j^\prime. 
\end{equation*}
Thus, 
\begin{equation*}
\begin{aligned}
\frac{\partial \mathcal{L}}{\partial \tilde{\vx}_n^\prime}\frac{\partial \mathcal{L}}{\partial \tilde{\vx}_m^\prime} &= \sum_{i,j=1}^{|\gB|}\frac{w_{ni} w_{mj}}{\gamma^2}(\sigma(\hat{\vy}_{ni}/\gamma)-\tilde{\vy}_{ni}) (\sigma(\hat{\vy}_{mj}/\gamma)-\tilde{\vy}_{mj})\tilde{\bm{\tau}}_i^{\prime\top}\tilde{\bm{\tau}}_j^\prime,
\end{aligned}
\end{equation*}
which can be rewritten as:
\begin{equation*}
\begin{aligned}
\frac{\partial \mathcal{L}}{\partial \tilde{\vx}_n^\prime}\frac{\partial \mathcal{L}}{\partial \tilde{\vx}_m^\prime} = &\sum_{i,j\neq n,m}^{|\gB|}\frac{w_{ni} w_{mj}}{\gamma^2}(\sigma(\hat{\vy}_{ni}/\gamma)-\tilde{\vy}_{ni}) (\sigma(\hat{\vy}_{mj}/\gamma)-\tilde{\vy}_{mj})\tilde{\bm{\tau}}_i^{\prime\top}\tilde{\bm{\tau}}_j^\prime\\
&+\sum_{i\neq n,m; j=n}^{|\gB|}\frac{w_{ni} w_{mn}}{\gamma^2}(\sigma(\hat{\vy}_{ni}/\gamma)-\tilde{\vy}_{ni}) (\sigma(\hat{\vy}_{mn}/\gamma)-\tilde{\vy}_{mn})\tilde{\bm{\tau}}_i^{\prime\top}\tilde{\bm{\tau}}_n^\prime \\
&+\sum_{i\neq n,m; j=m}^{|\gB|}\frac{w_{ni} w_{mm}}{\gamma^2}(\sigma(\hat{\vy}_{ni}/\gamma)-\tilde{\vy}_{ni}) (\sigma(\hat{\vy}_{mm}/\gamma)-\tilde{\vy}_{mm})\tilde{\bm{\tau}}_i^{\prime\top}\tilde{\bm{\tau}}_m^\prime \\
&+\sum_{i=n;j\neq n,m}^{|\gB|}\frac{w_{nn} w_{mj}}{\gamma^2}(\sigma(\hat{\vy}_{nn}/\gamma)-\tilde{\vy}_{nn}) (\sigma(\hat{\vy}_{mj}/\gamma)-\tilde{\vy}_{mj})\tilde{\bm{\tau}}_n^{\prime\top}\tilde{\bm{\tau}}_j^\prime\\
&+\sum_{i=m;j\neq n,m}^{|\gB|}\frac{w_{nm} w_{mj}}{\gamma^2}(\sigma(\hat{\vy}_{nm}/\gamma)-\tilde{\vy}_{nm}) (\sigma(\hat{\vy}_{mj}/\gamma)-\tilde{\vy}_{mj})\tilde{\bm{\tau}}_m^{\prime\top}\tilde{\bm{\tau}}_j^\prime \\
&+\frac{w_{nn} w_{mn}}{\gamma^2}(\sigma(\hat{\vy}_{nn}/\gamma)-\tilde{\vy}_{nn}) (\sigma(\hat{\vy}_{mn}/\gamma)-\tilde{\vy}_{mn})\tilde{\bm{\tau}}_n^{\prime\top}\tilde{\bm{\tau}}_n^\prime \\
&+\frac{w_{nn} w_{mm}}{\gamma^2}(\sigma(\hat{\vy}_{nn}/\gamma)-\tilde{\vy}_{nn}) (\sigma(\hat{\vy}_{mm}/\gamma)-\tilde{\vy}_{mm})\tilde{\bm{\tau}}_n^{\prime\top}\tilde{\bm{\tau}}_m^\prime \\
&+\frac{w_{nm} w_{mn}}{\gamma^2}(\sigma(\hat{\vy}_{nm}/\gamma)-\tilde{\vy}_{nm}) (\sigma(\hat{\vy}_{mn}/\gamma)-\tilde{\vy}_{mn})\tilde{\bm{\tau}}_m^{\prime\top}\tilde{\bm{\tau}}_n^\prime \\
&+\frac{w_{nm} w_{mm}}{\gamma^2}(\sigma(\hat{\vy}_{nm}/\gamma)-\tilde{\vy}_{nm}) (\sigma(\hat{\vy}_{mm}/\gamma)-\tilde{\vy}_{mm})\tilde{\bm{\tau}}_m^{\prime\top}\tilde{\bm{\tau}}_m^\prime .
\end{aligned}
\end{equation*}
In high-dimensional embedding spaces, both intra-modal and inter-modal negative pairs tend to be mutually orthogonal. Specifically, for any negative pair $(i, j)$, where $i \ne j$, 
\[
\tilde{\bm{\tau}}_i^{\prime\top} \tilde{\bm{\tau}}_j^\prime \approx 0.
\]
In our case, all pairs beyond $(i,j) \in \{(m,n), (n,m), (i,i)\}$ are negatives, thus we have,
\begin{equation*}
\begin{aligned}
\frac{\partial \mathcal{L}}{\partial \tilde{\vx}_n^\prime}\frac{\partial \mathcal{L}}{\partial \tilde{\vx}_m^\prime} \approx &\sum_{i=1}^{|\gB|}\frac{w_{ni} w_{mi}}{\gamma^2}(\sigma(\hat{\vy}_{ni}/\gamma)-\tilde{\vy}_{ni}) (\sigma(\hat{\vy}_{mi}/\gamma)-\tilde{\vy}_{mi})\tilde{\bm{\tau}}_i^{\prime\top}\tilde{\bm{\tau}}_i^\prime\\
&+\frac{w_{nn} w_{mm}}{\gamma^2}(\sigma(\hat{\vy}_{nn}/\gamma)-\tilde{\vy}_{nn}) (\sigma(\hat{\vy}_{mm}/\gamma)-\tilde{\vy}_{mm})\tilde{\bm{\tau}}_n^{\prime\top}\tilde{\bm{\tau}}_m^\prime \\
&+\frac{w_{nm} w_{mn}}{\gamma^2}(\sigma(\hat{\vy}_{nm}/\gamma)-\tilde{\vy}_{nm}) (\sigma(\hat{\vy}_{mn}/\gamma)-\tilde{\vy}_{mn})\tilde{\bm{\tau}}_m^{\prime\top}\tilde{\bm{\tau}}_n^\prime .
\end{aligned}
\end{equation*}
Because $(n,n)$ and $(m,m)$ are strictly aligned pairs, we have $\sigma(\hat{\vy}_{nn}/\gamma) \approx \tilde{\vy}_{nn} \approx 1$ and $\sigma(\hat{\vy}_{mm}/\gamma) \approx \tilde{\vy}_{mm} \approx 1$, hence $\sigma(\hat{\vy}_{nn}/\gamma) - \tilde{\vy}_{nn}$ and $\sigma(\hat{\vy}_{mm}/\gamma) - \tilde{\vy}_{mm}$ are close to zero. Therefore, we have:
\begin{equation*}
\begin{aligned}
\frac{\partial \mathcal{L}}{\partial \tilde{\vx}_n^\prime}\frac{\partial \mathcal{L}}{\partial \tilde{\vx}_m^\prime} \approx &\sum_{i\neq n,m}^{|\gB|}\frac{w_{ni} w_{mi}}{\gamma^2}(\sigma(\hat{\vy}_{ni}/\gamma)-\tilde{\vy}_{ni}) (\sigma(\hat{\vy}_{mi}/\gamma)-\tilde{\vy}_{mi})\tilde{\bm{\tau}}_i^{\prime\top}\tilde{\bm{\tau}}_i^\prime\\
&+\frac{w_{nm} w_{mn}}{\gamma^2}(\sigma(\hat{\vy}_{nm}/\gamma)-\tilde{\vy}_{nm}) (\sigma(\hat{\vy}_{mn}/\gamma)-\tilde{\vy}_{mn})\tilde{\bm{\tau}}_m^{\prime\top}\tilde{\bm{\tau}}_n^\prime. 
\end{aligned}
\end{equation*}
The first term captures the aggregated influence of shared negative examples on both \(\tilde{\vx}_n^\prime\) and \(\tilde{\vx}_m^\prime\), which affect them similarly and thus contribute little to their relative update direction.
In contrast, the second term reflects their mutual interaction and plays a dominant role in determining their representational divergence or alignment.

\section{Calculation of Concentration Ratio (\texttt{CR})}\label{app:cr}
To compute the \textit{concentration ratio} (\texttt{CR}), we use the surface area of a hyperspherical cap on the unit $(d{-}1)$-sphere, where $d$ is the dimensionality of the embedding space.  
Given a normalized cosine similarity value $c \in [0, 1]$, we consider the set of all unit vectors that form this similarity with a fixed reference direction.  
These vectors define a hyperspherical cap, a region on the surface of the unit hypersphere bounded by a fixed similarity threshold. The surface area ratio of this cap is given by:
\begin{equation*}
A = \mathcal{I}_{1 - c^2} \left( \frac{d - 1}{2}, \frac{1}{2} \right).
\end{equation*}
Here, $\mathcal{I}_x(a, b)$ denotes the regularized incomplete Beta function, defined as:
\begin{equation*}
\mathcal{I}_x(a, b) = \frac{ \int_0^x t^{a-1}(1 - t)^{b-1} \, dt }{ \int_0^1 t^{a-1}(1 - t)^{b-1} \, dt }.
\end{equation*}
This function describes the cumulative distribution of the Beta distribution and is widely used in geometric probability.  
In our context, it measures the proportion of the unit hypersphere’s surface that lies within a given angular range, equivalently, within a given cosine similarity of a fixed direction.  
Specifically, when computing hyperspherical cap areas, the variable substitution $x = 1 - c^2$ arises naturally from the spherical-to-cartesian coordinate transformation. 

We then define the concentration ratio as the complement of this surface ratio:
\begin{equation*}
\texttt{CR} = 1 - A .
\end{equation*}
This value reflects the proportion of the hypersphere surface that lies outside the similarity-defined cone.  
A higher \texttt{CR} indicates that the given similarity corresponds to a narrower directional region on the hypersphere, implying stronger feature concentration in the high-dimensional embedding space.

In implementation, we compute this value using the \texttt{scipy.special.betainc} function in Python.

\section{Implementation of Representation Blending}
\begin{algorithm}[H] 
\caption{$\operatorname{RepBlend}$:Representation Blending}
\label{alg:RepBlend}
\begin{algorithmic}[1]
\Require image and text representation $\{f^\text{imgE}(\tilde{\vx}_b), \tilde{\bm{\tau}}_b\}_{b=1}^{\gB}$ of one batch, Parameter $\alpha$ for $\operatorname{MixUP}$
	\Function{RepBlend}{$\{f^\text{imgE}(\tilde{\vx}_b), \tilde{\bm{\tau}}_b\}_{b=1}^{\gB}, \alpha$}
    \State $\{f^\text{imgE}(\tilde{\vx}_b)^{\text{shuf}}, \tilde{\bm{\tau}}_b^{\text{shuf}}\}_{b=1}^{\gB} \gets \text{shuffle}\big(\{f^\text{imgE}(\tilde{\vx}_b), \tilde{\bm{\tau}}_b\}_{b=1}^{\gB} \big)$ 
    \State $\triangleright$ \texttt{Shuffle image and text representations in one batch}
    \State Sample $\lambda$ from $\text{Beta}(\alpha, \alpha)$ for the batch
    \For{$b = 1$ to $|\gB|$}
    \State $\triangleright$ \texttt{Linear interpolation in representation space}
        \State $f^\text{imgE}(\tilde{\vx}_b) \gets \lambda f^\text{imgE}(\tilde{\vx}_b) + (1 - \lambda) f^\text{imgE}(\tilde{\vx}_b)^{\text{shuf}}$
        \State $\tilde{\bm{\tau}}_b \gets \lambda \tilde{\bm{\tau}}_b + (1 - \lambda) \tilde{\bm{\tau}}_b^{\text{shuf}}$ 
    \EndFor
    \Return $\{f^\text{imgE}(\tilde{\vx}_b), \tilde{\bm{\tau}}_b\}_{b=1}^{\gB}$
\EndFunction
	\end{algorithmic}
\end{algorithm}
\section{Comparison Methods}\label{app:comparison_methods}
\noindent{\textbf{Coreset Selection Methods.}} 

1) Random (Rand): Randomly selects a subset of samples from the full dataset to form a coreset. While this approach is unbiased, it may fail to capture the most informative or representative instances necessary for efficient training.

2) Herding (Herd)~\citep{c-herd}: Selects samples based on herding dynamics to approximate the mean of the data distribution. It iteratively chooses instances that minimize the discrepancy between the coreset and the full dataset’s feature distribution.

3) K-Center (K-Cent)~\citep{c-kcenter}: Selects samples that serve as representative centers in the feature space. It aims to maximize coverage by iteratively choosing points that are maximally distant from the already selected ones.

4) Forgetting (Forget)~\citep{c-forget}: Selects samples based on how often they are forgotten during training, i.e., when correct predictions become incorrect. Samples with low forgetting counts are removed first, prioritizing the retention of harder and more informative examples.

\noindent{\textbf{Dataset Distillation Methods.}} 

1) MTT-VL~\cite{wu2024visionlanguage}: The first MDD approach that extends the trajectory matching framework MTT~\citep{mtt} to vision-language data, enabling dataset distillation in multimodal settings.

2) TESLA-VL~\cite{xu2024lors}: An efficient variant of the MTT framework, TESLA~\cite{tesla}, implemented in LoRS~\cite{xu2024lors} as an ablation to evaluate the effectiveness of similarity mining in multimodal distillation.

3) LoRS~\citep{xu2024lors}: A sota MDD method that distills both image-text pairs and their similarity matrix to enhance multimodal distillation, while leveraging low-rank factorization for improving efficiency.
\section{Hyperparameter Settings}\label{app:hyperparameter_settings}
The hyperparameter settings, summarized in \autoref{tab:Hyperparameter_buffer} and \autoref{tab:Hyperparameter_distillation}, follow the configurations used in LoRS~\cite{xu2024lors} to ensure fair and consistent comparisons.
\begin{table}[t]
\vspace{-1em}
    \caption{Hyperparameter settings for buffer.}
    \centering
    \small
    \begin{tabular}{l|cc}
        \toprule
        & Flickr-30K & MS-COCO\\ \midrule
        epoch & 10 & 10\\
num\_experts & 20& 20 \\
batch\_size & 128& 128 \\
lr\_teacher\_img & 0.1& 0.1 \\
lr\_teacher\_txt & 0.1& 0.1 \\
image\_size & 224$\times$224& 224$\times$224 \\
        \bottomrule
    \end{tabular}
    \label{tab:Hyperparameter_buffer}
\end{table}
\begin{table}[t]
    \caption{Hyperparameter settings for distillation.}
    \centering
    \small
    \begin{tabular}{l|ccc|ccc}
        \toprule
        & \multicolumn{3}{c}{Flickr-30K} & \multicolumn{3}{c}{MS-COCO}\\ 
        & 100 pairs & 200 pairs & 500 pairs & 100 pairs & 200 pairs & 500 pairs\\ \midrule
        syn\_steps & 8 & 8 & 8 & 8 & 8 & 8 \\
        expert\_epochs & 1 & 1 & 1 & 1 & 1 & 1 \\
        max\_start\_epoch & 2 & 2 & 3 & 2 & 2 & 2 \\
        iteration & 2000 & 2000 & 2000 & 2000 & 2000 & 2000 \\
        lr\_img & 100 & 1000 & 1000 & 1000 & 1000 & 5000 \\
        lr\_txt & 100 & 1000 & 1000 & 1000 & 1000 & 5000 \\
        lr\_lr & 1e-2 & 1e-2 & 1e-2 & 1e-2 & 1e-2 & 1e-2 \\
        lr\_teacher\_img & 0.1 & 0.1 & 0.1 & 0.1 & 0.1 & 0.1 \\
        lr\_teacher\_txt & 0.1 & 0.1 & 0.1 & 0.1 & 0.1 & 0.1 \\
        lr\_sim & 10.0 & 10.0 & 100.0 & 5.0 & 50.0 & 500.0 \\
        sim\_type & lowrank & lowrank & lowrank & lowrank & lowrank & lowrank \\
        sim\_rank & 10 & 5 & 20 & 10 & 20 & 40 \\
        sim\_alpha & 3.0 & 1.0 & 0.01 & 1.0 & 1.0 & 1.0 \\ 
        num\_queries & 99 & 199 & 499 & 99 & 199 & 499 \\
        mini\_batch\_size & 20 & 20 & 40 & 20 & 20 & 30 \\
        loss\_type & WBCE & WBCE & WBCE & WBCE & WBCE & WBCE \\
        beta\_distribution & $\alpha=1.0$ & $\alpha=1.0$ & $\alpha=1.0$ & $\alpha=1.0$ & $\alpha=1.0$ & $\alpha=1.0$ \\
        \bottomrule
    \end{tabular}
    \label{tab:Hyperparameter_distillation}
\end{table}

\textcolor{black}{\section{Generalization to Audio-Text Datasets}\label{app:generalization_to_audio_text}}

To explore the generalizability of our multimodal dataset distillation approach beyond image-text data, we extend our experiments to the audio-text domain using the AudioCaps~\cite{audiocaps} dataset. AudioCaps is a widely used dataset for audio-text contrastive learning, derived from AudioSet~\cite{audioset}. It comprises approximately 44,000 audio clips paired with human-annotated captions that vividly describe the auditory content. The distillation process follows a similar protocol to that used in the image-text experiments. We employ BERT as the text encoder and EfficientAT (mn20\_as)~\cite{schmid2023efficient} as the audio encoder. EfficientAT is a state-of-the-art audio classification model based on MobileNet~\cite{mobilenet}, designed to achieve high representational quality with low computational overhead.

The results presented in ~\autoref{tab:audiocaps}, compare our method against LoRS~\cite{xu2024lors} on the AudioCaps dataset for 100, 200, and 500 synthetic pairs. Our approach consistently outperforms LoRS across all metrics and data scales. In 500 pairs settings, our method achieves AR@10 of 46.8 and TR@10 of 54.1, compared to LoRS's 36.7 and 41.3, respectively. Notably, our method achieves around 65\% of the full dataset's performance using only 1.13\% of the data. Superior results demonstrate that our proposed approach successfully generalizes to audio-text datasets, extending beyond the image-text domain. By achieving significant performance gains over existing baseline, our method establishes a more robust framework for multimodal dataset distillation across diverse modality pairs.

\begin{table}[h]
\centering
\caption{Results on AudioCaps~\cite{audiocaps}. Both distillation and validation are performed using pretrained EfficientAT+BERT. The model trained on full dataset performs: AR@1=21.3, AR@5=53.2, AR@10=68.5; TR@1=25.2, TR@5=58.8, TR@10=71.6.}
\label{tab:audiocaps}
\resizebox{0.97\textwidth}{!}{%
    \begin{tabular}{lcccccccccc}
    \toprule
    \multirow{2}{*}{Method} & \multirow{2}{*}{Pairs} & \multirow{2}{*}{Ratio} & \multicolumn{3}{c}{Audio to Text} & \multicolumn{3}{c}{Text to Audio} \\
    \cmidrule(lr){4-6} \cmidrule(lr){7-9}
    & & & {AR@1} & {AR@5} & {AR@10} & {TR@1} & {TR@5} & {TR@10} \\
    \midrule
    \multirow{3}{*}{LoRS~\cite{xu2024lors}} 
    & 100 & 0.23\% & 2.7$\pm$0.3 & 8.6$\pm$0.3 & 14.7$\pm$0.4 & 5.9$\pm$0.3 & 13.0$\pm$0.4 & 21.8$\pm$0.5 \\
    & 200 & 0.45\% & 3.8$\pm$0.2 & 14.8$\pm$0.2 & 21.8$\pm$0.2 & 8.0$\pm$0.2 & 21.2$\pm$0.2 & 33.1$\pm$0.2 \\
    & 500 & 1.13\% & 7.1$\pm$0.1 & 24.7$\pm$0.2 & 36.7$\pm$0.2 & 9.2$\pm$0.2 & 27.4$\pm$0.3 & 41.3$\pm$0.3 \\
    \midrule
    \multirow{3}{*}{Ours} 
    & 100 & 0.23\% & 4.1$\pm$0.2 & 14.2$\pm$0.3 & 23.7$\pm$0.4 & 8.9$\pm$0.1 & 24.3$\pm$0.2 & 34.7$\pm$0.3 \\
    & 200 & 0.45\% & 6.8$\pm$0.2 & 20.6$\pm$0.2 & 31.4$\pm$0.3 & 9.7$\pm$0.2 & 29.1$\pm$0.4 & 41.2$\pm$0.4 \\
    & 500 & 1.13\% & 9.7$\pm$0.1 & 32.2$\pm$0.3 & 46.8$\pm$0.2 & 13.8$\pm$0.3 & 38.6$\pm$0.3 & 54.1$\pm$0.4 \\
    \bottomrule
    \end{tabular}%
    }
\end{table}

\section{More Experiments on Various Architectures}\label{app:multi_arch}
\begin{figure}
    \centering
    \includegraphics[width=1\linewidth]{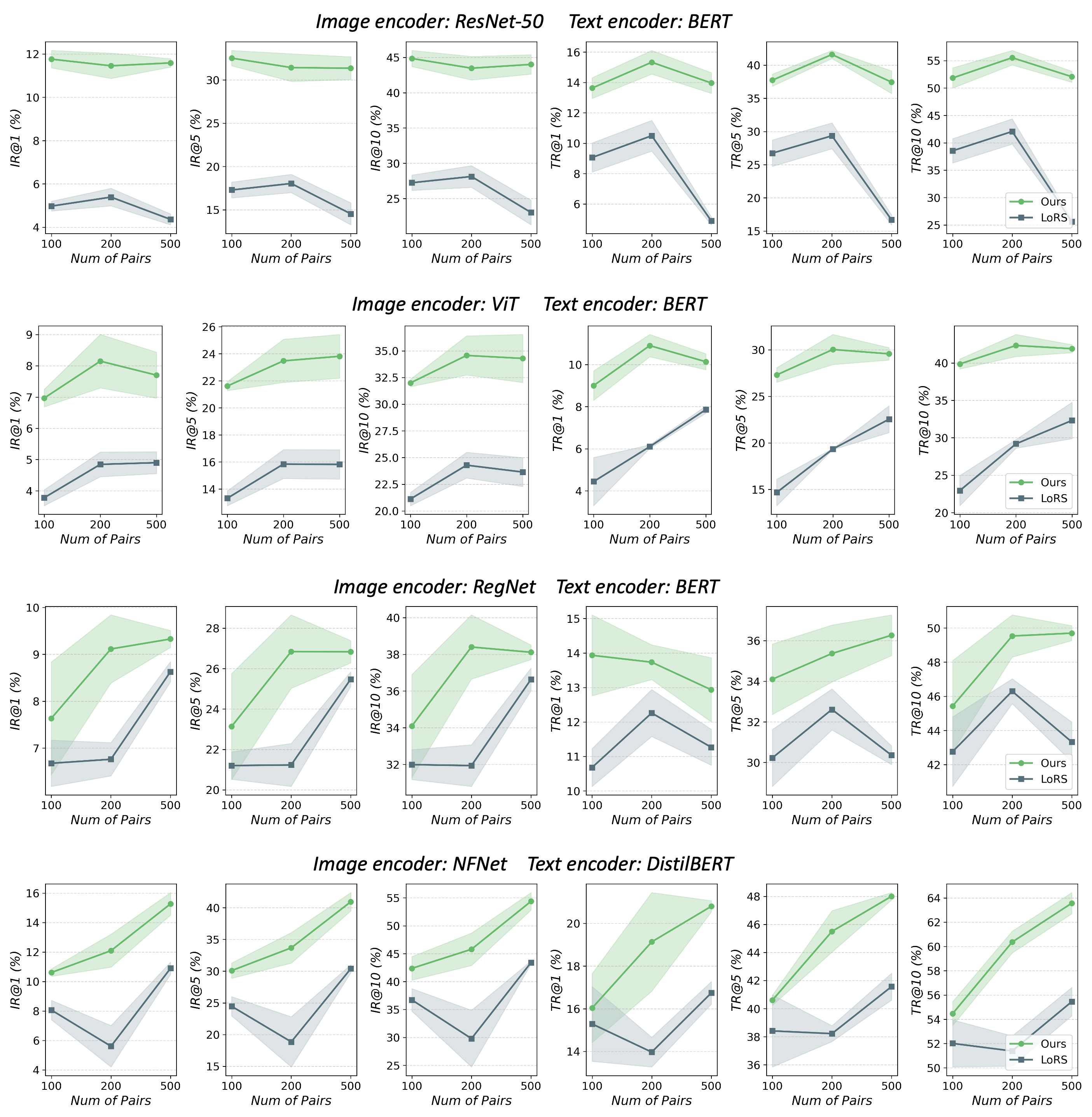}
    \caption{Performance on Flickr-30K with different combinations of image and text encoders.}
    \label{fig:multiarch_Flickr}
\end{figure}

\begin{figure}
    \centering
    \includegraphics[width=1\linewidth]{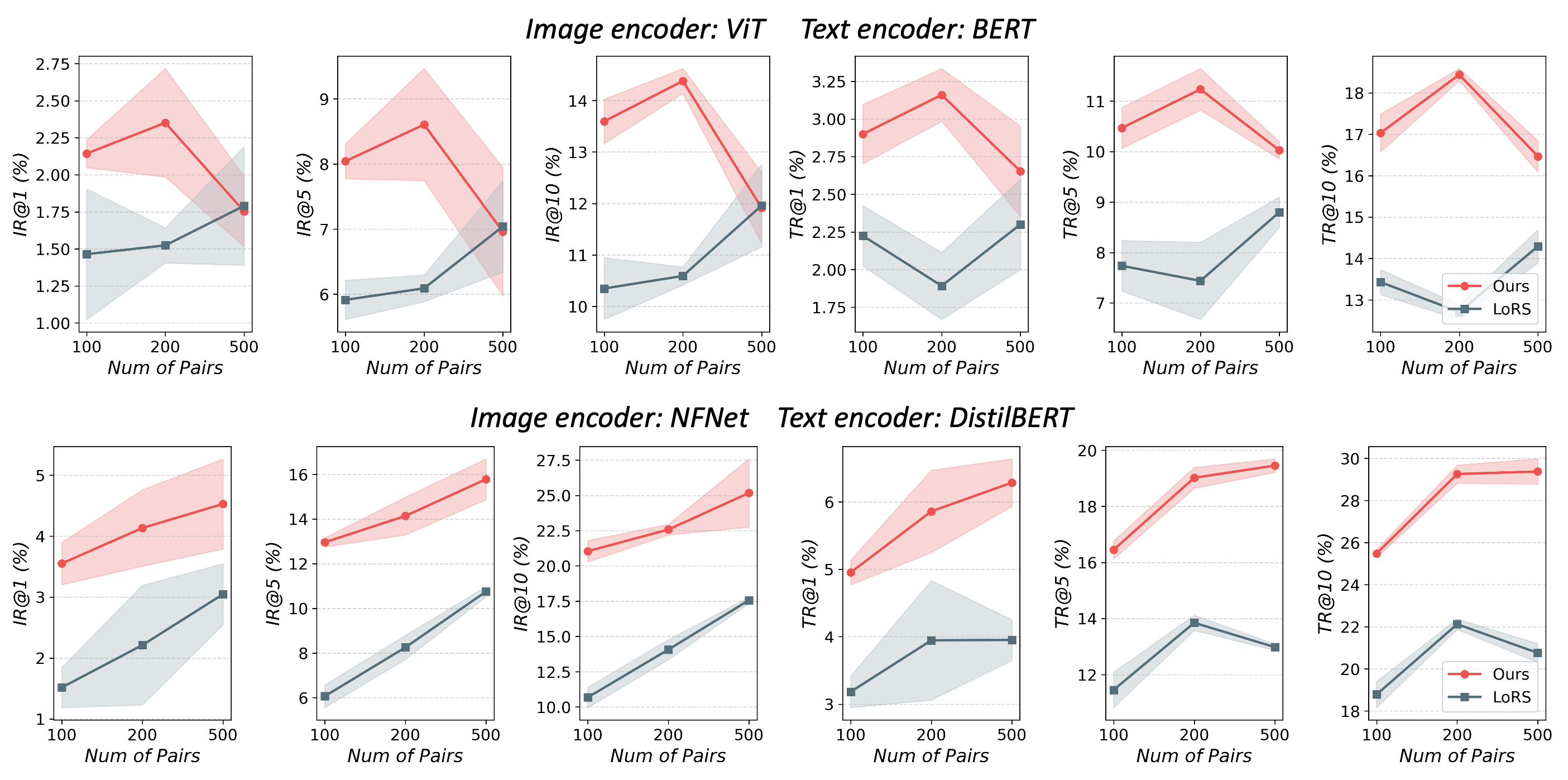}
    \caption{Performance on MS-COCO with different combinations of image and text encoders.}
    \label{fig:multiarch_coco}
\end{figure}

We further implement our method using different combinations of image encoders (e.g., ResNet-50~\cite{resnet}, ViT~\cite{vit}, RegNet~\cite{regnet}, NFNet~\cite{brock2021high}) and text encoders (e.g., BERT~\cite{devlin2018bert}, DistilBERT~\cite{sanh2019distilbert}) to assess the robustness and generality of our framework. The corresponding results are presented in \autoref{fig:multiarch_Flickr} and \autoref{fig:multiarch_coco}. Across all architecture combinations, our method consistently outperforms the baseline LoRS~\cite{xu2024lors}, demonstrating its adaptability to various vision and language backbones. 

\section{Compare with SRe2L~\cite{sre} on Classification Tasks}
To assess our dataset distillation method for classification task under low IPC\footnote{IPC denotes image per class} settings, we experimented on ImageNet-100, a 100-class subset of ImageNet-1K~\cite{imagenet}. We compared our method against SRe2L~\cite{sre}, a leading distillation approach. As our method focuses on multimodal distillation, we assigned uniform text descriptions (“A picture of [ClassName]”) to images of the same class. During evaluation, test images and class descriptions were processed by image and text branches to generate embeddings, with classification based on the highest similarity score. For SRe2L~\cite{sre}, we followed its original setup, recovering data from ResNet-18 trained for 100 epochs on the full dataset, using 4k recovery iterations and a softmax temperature of 20.

\begin{figure}[ht]
    \centering
    \def\imgwidth{2.5cm}
    \def\imgheight{2.5cm}
    \def\colspace{0.2cm} 
    \def\rowspace{0.2cm} 
    \def\labelspace{0.3cm} 
    \def\groupspace{0.3cm} 
    
    \begin{tikzpicture}
        \def\labelA{White shark}
        \def\labelB{Electric ray}
        \def\labelC{Bulbul}
        \def\labelD{Banded gecko}
        \def\labelE{Whiptail}
        
        \matrix (firstgroup) [
            matrix of nodes,
            nodes={inner sep=0pt, outer sep=0pt},
            column sep=\colspace,
            row sep=\rowspace
        ] {
            \includegraphics[width=\imgwidth,height=\imgheight]{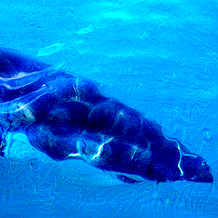} &
            \includegraphics[width=\imgwidth,height=\imgheight]{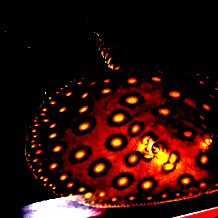} &
            \includegraphics[width=\imgwidth,height=\imgheight]{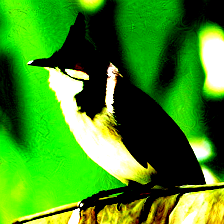} &
            \includegraphics[width=\imgwidth,height=\imgheight]{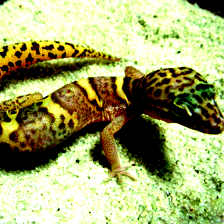} &
            \includegraphics[width=\imgwidth,height=\imgheight]{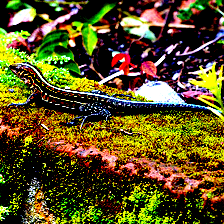} \\
            \includegraphics[width=\imgwidth,height=\imgheight]{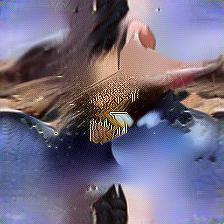} &
            \includegraphics[width=\imgwidth,height=\imgheight]{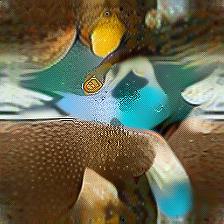} &
            \includegraphics[width=\imgwidth,height=\imgheight]{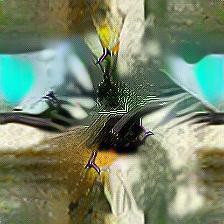} &
            \includegraphics[width=\imgwidth,height=\imgheight]{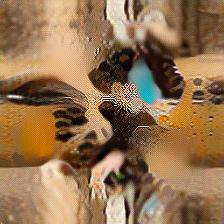} &
            \includegraphics[width=\imgwidth,height=\imgheight]{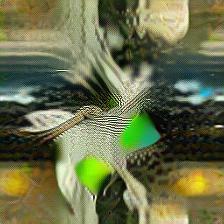} \\
        };
        
        \node[font=\sffamily, rotate=90, anchor=center] at ([xshift=-0.3cm]firstgroup-1-1.west) {\textbf{Ours}};
        \node[font=\sffamily, rotate=90, anchor=center] at ([xshift=-0.3cm]firstgroup-2-1.west) {\textbf{SRe2L}};
        
        \node[font=\sffamily\small, align=center, text width=\imgwidth] at ([yshift=-\labelspace]firstgroup-2-1.south) {\labelA};
        \node[font=\sffamily\small, align=center, text width=\imgwidth] at ([yshift=-\labelspace]firstgroup-2-2.south) {\labelB};
        \node[font=\sffamily\small, align=center, text width=\imgwidth] at ([yshift=-\labelspace]firstgroup-2-3.south) {\labelC};
        \node[font=\sffamily\small, align=center, text width=\imgwidth] at ([yshift=-\labelspace]firstgroup-2-4.south) {\labelD};
        \node[font=\sffamily\small, align=center, text width=\imgwidth] at ([yshift=-\labelspace]firstgroup-2-5.south) {\labelE};
    \end{tikzpicture}
    
    \vspace{\groupspace}
    
    \begin{tikzpicture}
        \def\labelF{Triceratops}
        \def\labelG{Hognose snake}
        \def\labelH{Garden spider}
        \def\labelI{Bee-eater}
        \def\labelJ{Hornbill}
        
        \matrix (secondgroup) [
            matrix of nodes,
            nodes={inner sep=0pt, outer sep=0pt},
            column sep=\colspace,
            row sep=\rowspace
        ] {
            \includegraphics[width=\imgwidth,height=\imgheight]{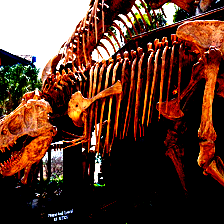} &
            \includegraphics[width=\imgwidth,height=\imgheight]{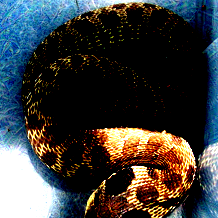} &
            \includegraphics[width=\imgwidth,height=\imgheight]{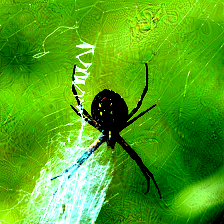} &
            \includegraphics[width=\imgwidth,height=\imgheight]{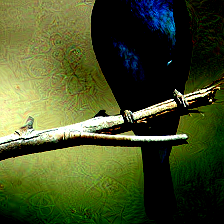} &
            \includegraphics[width=\imgwidth,height=\imgheight]{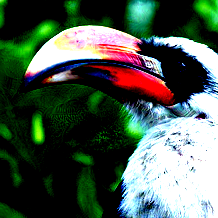} \\
            \includegraphics[width=\imgwidth,height=\imgheight]{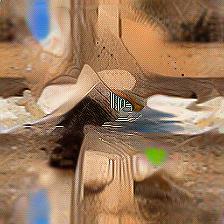} &
            \includegraphics[width=\imgwidth,height=\imgheight]{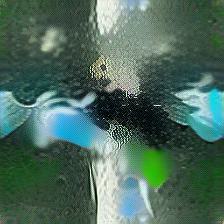} &
            \includegraphics[width=\imgwidth,height=\imgheight]{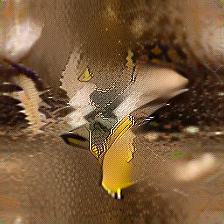} &
            \includegraphics[width=\imgwidth,height=\imgheight]{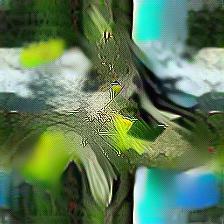} &
            \includegraphics[width=\imgwidth,height=\imgheight]{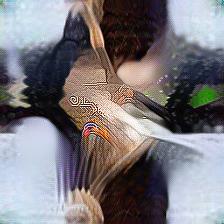} \\
        };
        
        \node[font=\sffamily, rotate=90, anchor=center] at ([xshift=-0.3cm]secondgroup-1-1.west) {\textbf{Ours}};
        \node[font=\sffamily, rotate=90, anchor=center] at ([xshift=-0.3cm]secondgroup-2-1.west) {\textbf{SRe2L}};
        
        \node[font=\sffamily\small, align=center, text width=\imgwidth] at ([yshift=-\labelspace]secondgroup-2-1.south) {\labelF};
        \node[font=\sffamily\small, align=center, text width=\imgwidth] at ([yshift=-\labelspace]secondgroup-2-2.south) {\labelG};
        \node[font=\sffamily\small, align=center, text width=\imgwidth] at ([yshift=-\labelspace]secondgroup-2-3.south) {\labelH};
        \node[font=\sffamily\small, align=center, text width=\imgwidth] at ([yshift=-\labelspace]secondgroup-2-4.south) {\labelI};
        \node[font=\sffamily\small, align=center, text width=\imgwidth] at ([yshift=-\labelspace]secondgroup-2-5.south) {\labelJ};
    \end{tikzpicture}
    
    \caption{Synthetic data visualization on ImageNet-100 from our approach and SRe2L when IPC=1.}
    \label{fig:visualization_imgnet}
\end{figure}

\autoref{tab:sre2l} shows our method significantly outperforms SRe2L on ImageNet-100 at low IPC. Results were obtained by training models from scratch on distilled data and testing on the test set. At \texttt{ipc}=1, our method achieves 65.8\% Top-1 and 89.9\% Top-5 accuracy, compared to SRe2L’s 2.5\% and 9.2\%. This significant improvement can be attributed to the inclusion of a text projection head, distilled text embeddings, and the learned similarity matrix. Meanwhile, this added complexity is on par with the soft label augmentation used in SRe2L.
Some visualization of distilled data are shown in \autoref{fig:visualization_imgnet}.

\begin{table}[h]
\centering
\caption{Comparison of Our Method with SRe2L~\cite{sre} on ImageNet-100 Dataset}
\label{tab:sre2l}
\begin{tabular}{lcccc}
\toprule
\multirow{2}{*}{{\texttt{ipc}}} & \multicolumn{2}{c}{{Ours}} & \multicolumn{2}{c}{{SRe2L~\cite{sre}}} \\
\cmidrule(lr){2-3} \cmidrule(lr){4-5}
& {Acc1} & {Acc5} & {Acc1} & {Acc5} \\
\midrule
{1} & 65.8$\pm$0.2 & 89.9$\pm$0.5 & 2.5$\pm$0.2 & 9.2$\pm$0.3 \\
\bottomrule
\end{tabular}
\end{table}

\section{Visualization of Distilled Data}\label{app:visualization}
\begin{figure}

    \centering
    \def\imgwidth{3.1cm}
    \def\imgheight{3.1cm}
    \def\txtwidth{3.1cm}
    \def\innersep{1.1pt}
    \def\groupsep{0.2cm}
    \def\borderwidth{2.5pt}
    \begin{tikzpicture}
    
    \matrix [column sep=0.01cm, row sep=0cm] {
        \node[draw=blue!50, rounded corners, inner sep=\innersep, line width=\borderwidth] {
            \includegraphics[width=\imgwidth,height=\imgheight]{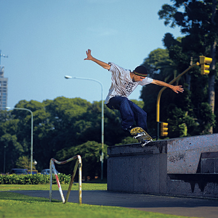}
        };
        &
        \node[draw=red!50, rounded corners, inner sep=\innersep, line width=\borderwidth] {
            \includegraphics[width=\imgwidth,height=\imgheight]{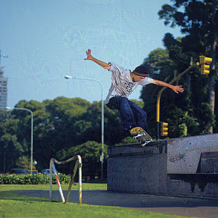}
        };
        &[\groupsep]
        \node[draw=blue!50, rounded corners, inner sep=\innersep, line width=\borderwidth] {
            \includegraphics[width=\imgwidth,height=\imgheight]{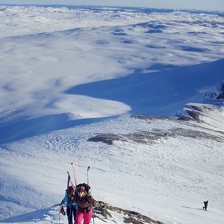}
        };
        &
        \node[draw=red!50, rounded corners, inner sep=\innersep, line width=\borderwidth] {
            \includegraphics[width=\imgwidth,height=\imgheight]{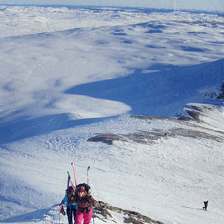}
        };
        %
        \\
        
        \node[text width=\txtwidth, align=center, text=blue!50, font=\rmfamily\bfseries\small] {
           a boy in a white t-shirt does skateboard tricks
        };
        &
        \node[text width=\txtwidth, align=center, text=red!50, font=\rmfamily\bfseries\small] {
            a skateboarder does a tailslide down the side of a railing over some stairs
        };
        &[\groupsep]
        \node[text width=\txtwidth, align=center, text=blue!50, font=\rmfamily\bfseries\small] {
            the two people carrying skis are walking in the snow
        };
        &
        \node[text width=\txtwidth, align=center, text=red!50, font=\rmfamily\bfseries\small] {
            three skiers are standing on a snowy hilltop
        };
        \\
    };
    \end{tikzpicture}

    \begin{tikzpicture}
    
    \matrix [column sep=0.01cm, row sep=0cm] {
        \node[draw=blue!50, rounded corners, inner sep=\innersep, line width=\borderwidth] {
            \includegraphics[width=\imgwidth,height=\imgheight]{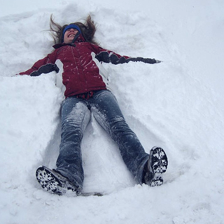}
        };
        &
        \node[draw=red!50, rounded corners, inner sep=\innersep, line width=\borderwidth] {
            \includegraphics[width=\imgwidth,height=\imgheight]{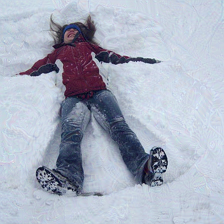}
        };
        &[\groupsep]
        \node[draw=blue!50, rounded corners, inner sep=\innersep, line width=\borderwidth] {
            \includegraphics[width=\imgwidth,height=\imgheight]{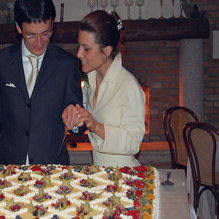}
        };
        &
        \node[draw=red!50, rounded corners, inner sep=\innersep, line width=\borderwidth] {
            \includegraphics[width=\imgwidth,height=\imgheight]{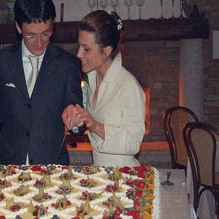}
        };
        \\
        
        \node[text width=\txtwidth, align=center, text=blue!50, font=\rmfamily\bfseries\small] {
           the young girl makes a wonderful snow angel on the ground
        };
        &
        \node[text width=\txtwidth, align=center, text=red!50, font=\rmfamily\bfseries\small] {
            a child wearing a gray winter coat and blue snow boots is unhappy upon discovering their playhouse covered in snow
        };
        &[\groupsep]
        \node[text width=\txtwidth, align=center, text=blue!50, font=\rmfamily\bfseries\small] {
            a man and a woman are at a restaurant cutting cake
        };
        &
        \node[text width=\txtwidth, align=center, text=red!50, font=\rmfamily\bfseries\small] {
            a woman wearing a burgundy shirt and a man wearing black prepare a tray of cups of wine
        };
        \\
    };
    \end{tikzpicture}

    \begin{tikzpicture}
    
    \matrix [column sep=0.01cm, row sep=0cm] {
        \node[draw=blue!50, rounded corners, inner sep=\innersep, line width=\borderwidth] {
            \includegraphics[width=\imgwidth,height=\imgheight]{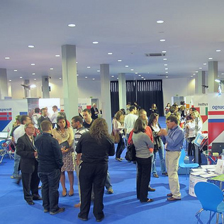}
        };
        &
        \node[draw=red!50, rounded corners, inner sep=\innersep, line width=\borderwidth] {
            \includegraphics[width=\imgwidth,height=\imgheight]{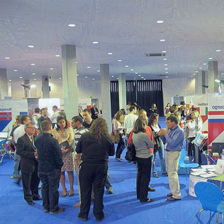}
        };
        &[\groupsep]
        \node[draw=blue!50, rounded corners, inner sep=\innersep, line width=\borderwidth] {
            \includegraphics[width=\imgwidth,height=\imgheight]{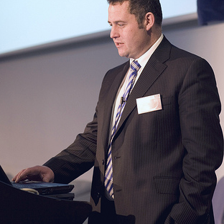}
        };
        &
        \node[draw=red!50, rounded corners, inner sep=\innersep, line width=\borderwidth] {
            \includegraphics[width=\imgwidth,height=\imgheight]{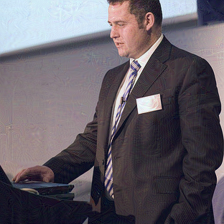}
        };
        \\
        
        \node[text width=\txtwidth, align=center, text=blue!50, font=\rmfamily\bfseries\small] {
            a group of people are gathered in an office, talking
        };
        &
        \node[text width=\txtwidth, align=center, text=red!50, font=\rmfamily\bfseries\small] {
            many people sit and socialize in groups inside a restaurant
        };
        &[\groupsep]
        \node[text width=\txtwidth, align=center, text=blue!50, font=\rmfamily\bfseries\small] {
           a man giving a presentation while wearing a brown striped suit
        };
        &
        \node[text width=\txtwidth, align=center, text=red!50, font=\rmfamily\bfseries\small] {
           dark-haired man with beard reading while drinking coffee at a counter
        };
        \\
    };
    \end{tikzpicture}

    \begin{tikzpicture}
    
    \matrix [column sep=0.01cm, row sep=0cm] {
        \node[draw=blue!50, rounded corners, inner sep=\innersep, line width=\borderwidth] {
            \includegraphics[width=\imgwidth,height=\imgheight]{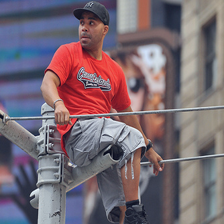}
        };
        &
        \node[draw=red!50, rounded corners, inner sep=\innersep, line width=\borderwidth] {
            \includegraphics[width=\imgwidth,height=\imgheight]{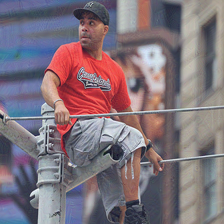}
        };
        &[\groupsep]
        \node[draw=blue!50, rounded corners, inner sep=\innersep, line width=\borderwidth] {
            \includegraphics[width=\imgwidth,height=\imgheight]{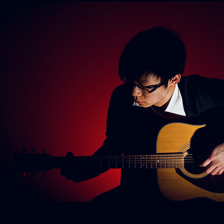}
        };
        &
        \node[draw=red!50, rounded corners, inner sep=\innersep, line width=\borderwidth] {
            \includegraphics[width=\imgwidth,height=\imgheight]{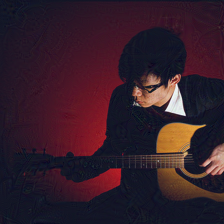}
        };
        \\
        
        \node[text width=\txtwidth, align=center, text=blue!50, font=\rmfamily\bfseries\small] {
            a guy on a city electric pole watching over
        };
        &
        \node[text width=\txtwidth, align=center, text=red!50, font=\rmfamily\bfseries\small] {
            a man in a yellow shirt and helmet hanging from a huge beam on ropes
        };
        &[\groupsep]
        \node[text width=\txtwidth, align=center, text=blue!50, font=\rmfamily\bfseries\small] {
            a young, asian man is seen playing his guitar
        };
        &
        \node[text width=\txtwidth, align=center, text=red!50, font=\rmfamily\bfseries\small] {
            man playing a brown and white electric guitar
        };
        \\
    };
    \end{tikzpicture}
    \caption{Flickr-30K before and after distillation. \textcolor{blue}{(}\textit{\textcolor{blue}{Left}}\textcolor{blue}{)} The original image-text pairs before the distillation. \textcolor{red}{(}\textit{\textcolor{red}{Right}}\textcolor{red}{)} The 
     image-text pairs after distillation. }
    \label{fig:visualization_flickr}
\end{figure}
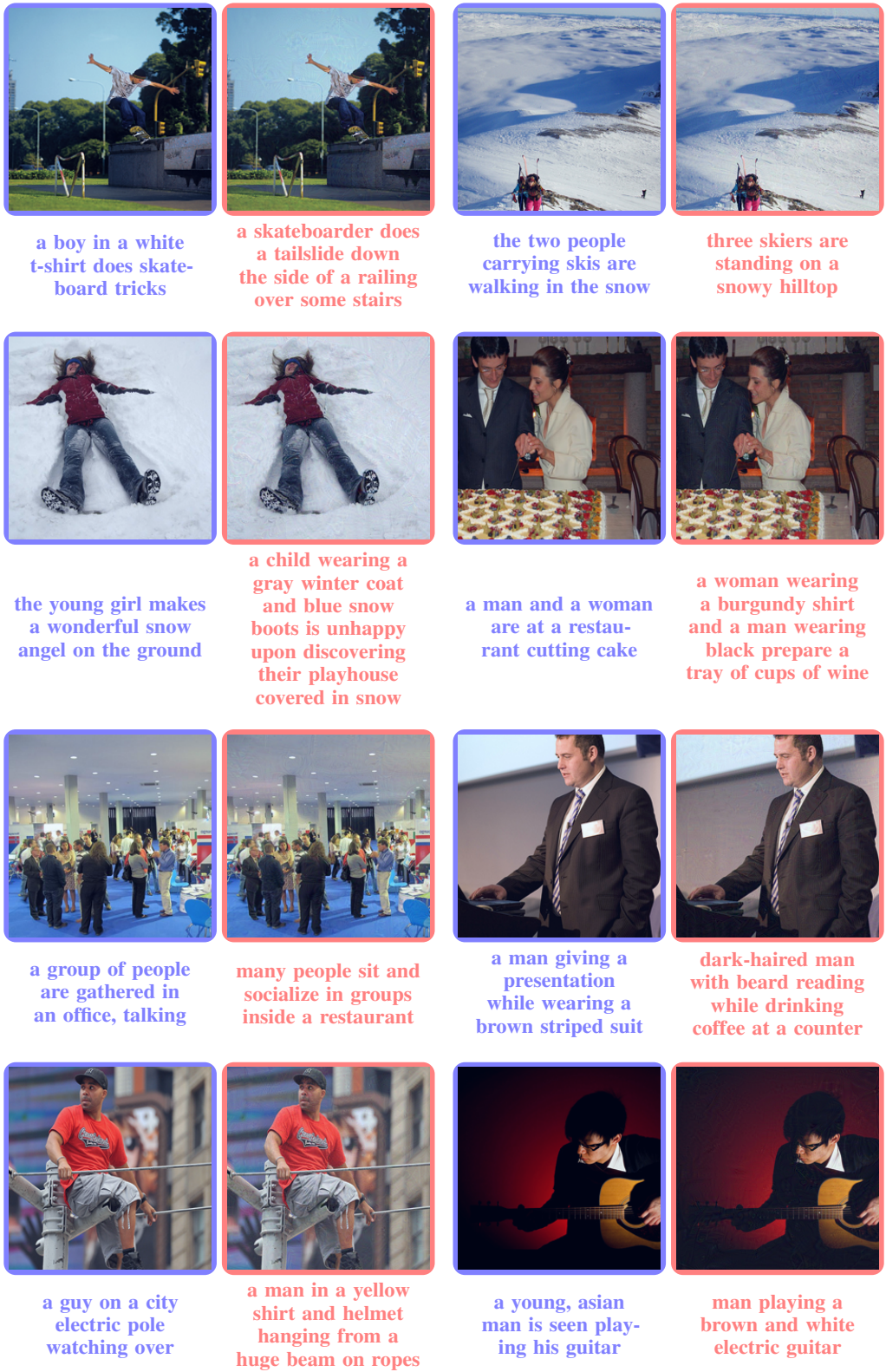

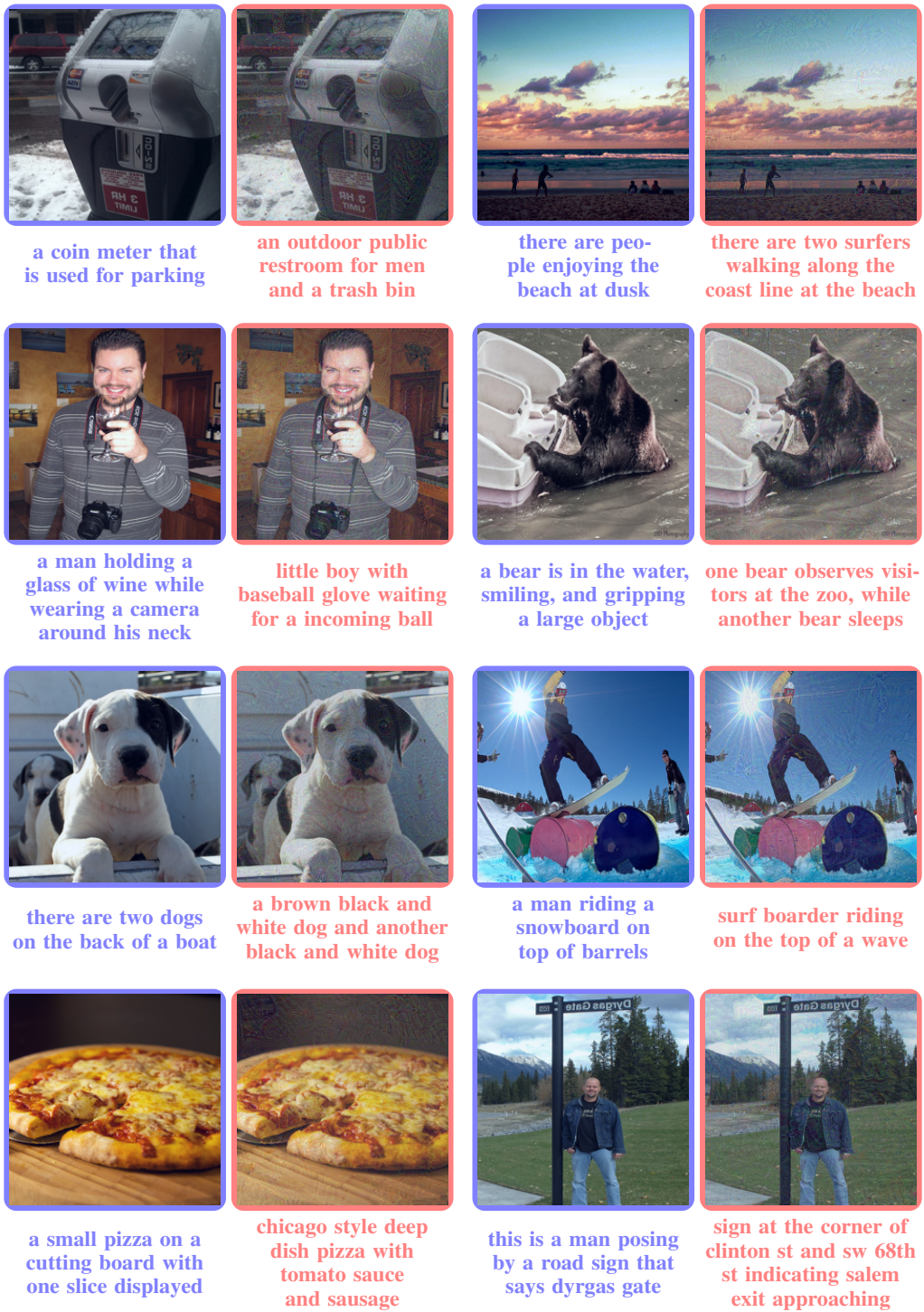
\begin{figure}

    \centering
    \def\imgwidth{3.1cm}
    \def\imgheight{3.1cm}
    \def\txtwidth{3.1cm}
    \def\innersep{1.1pt}
    \def\groupsep{0.2cm}
    \def\borderwidth{2.5pt}
    \begin{tikzpicture}
    
    \matrix [column sep=0.01cm, row sep=0cm] {
        \node[draw=blue!50, rounded corners, inner sep=\innersep, line width=\borderwidth] {
            \includegraphics[width=\imgwidth,height=\imgheight]{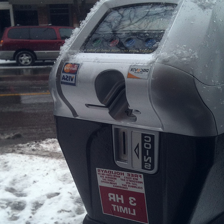}
        };
        &
        \node[draw=red!50, rounded corners, inner sep=\innersep, line width=\borderwidth] {
            \includegraphics[width=\imgwidth,height=\imgheight]{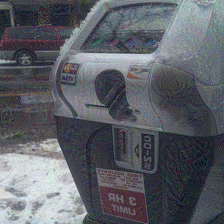}
        };
        &[\groupsep]
        \node[draw=blue!50, rounded corners, inner sep=\innersep, line width=\borderwidth] {
            \includegraphics[width=\imgwidth,height=\imgheight]{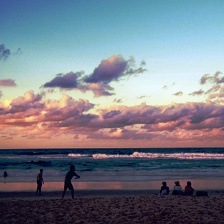}
        };
        &
        \node[draw=red!50, rounded corners, inner sep=\innersep, line width=\borderwidth] {
            \includegraphics[width=\imgwidth,height=\imgheight]{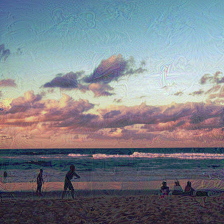}
        };
        %
        \\
        
        \node[text width=\txtwidth, align=center, text=blue!50, font=\rmfamily\bfseries\small] {
           a coin meter that is used for parking
        };
        &
        \node[text width=\txtwidth, align=center, text=red!50, font=\rmfamily\bfseries\small] {
            an outdoor public restroom for men and a trash bin
        };
        &[\groupsep]
        \node[text width=\txtwidth, align=center, text=blue!50, font=\rmfamily\bfseries\small] {
            there are people enjoying the beach at dusk
        };
        &
        \node[text width=\txtwidth, align=center, text=red!50, font=\rmfamily\bfseries\small] {
            there are two surfers walking along the coast line at the beach
        };
        \\
    };
    \end{tikzpicture}

    \begin{tikzpicture}
    
    \matrix [column sep=0.01cm, row sep=0cm] {
        \node[draw=blue!50, rounded corners, inner sep=\innersep, line width=\borderwidth] {
            \includegraphics[width=\imgwidth,height=\imgheight]{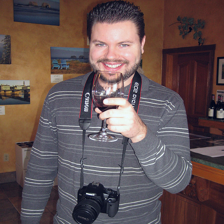}
        };
        &
        \node[draw=red!50, rounded corners, inner sep=\innersep, line width=\borderwidth] {
            \includegraphics[width=\imgwidth,height=\imgheight]{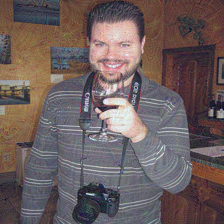}
        };
        &[\groupsep]
        \node[draw=blue!50, rounded corners, inner sep=\innersep, line width=\borderwidth] {
            \includegraphics[width=\imgwidth,height=\imgheight]{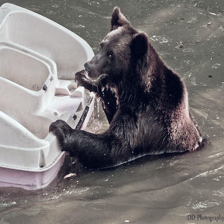}
        };
        &
        \node[draw=red!50, rounded corners, inner sep=\innersep, line width=\borderwidth] {
            \includegraphics[width=\imgwidth,height=\imgheight]{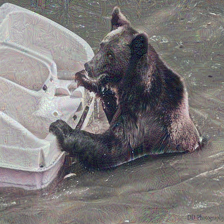}
        };
        \\
        
        \node[text width=\txtwidth, align=center, text=blue!50, font=\rmfamily\bfseries\small] {
           a man holding a glass of wine while wearing a camera around his neck
        };
        &
        \node[text width=\txtwidth, align=center, text=red!50, font=\rmfamily\bfseries\small] {
            little boy with baseball glove waiting for a incoming ball
        };
        &[\groupsep]
        \node[text width=\txtwidth, align=center, text=blue!50, font=\rmfamily\bfseries\small] {
            a bear is in the water, smiling, and gripping a large object
        };
        &
        \node[text width=\txtwidth, align=center, text=red!50, font=\rmfamily\bfseries\small] {
            one bear observes visitors at the zoo, while another bear sleeps
        };
        \\
    };
    \end{tikzpicture}

    \begin{tikzpicture}
    
    \matrix [column sep=0.01cm, row sep=0cm] {
        \node[draw=blue!50, rounded corners, inner sep=\innersep, line width=\borderwidth] {
            \includegraphics[width=\imgwidth,height=\imgheight]{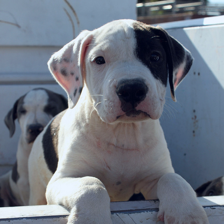}
        };
        &
        \node[draw=red!50, rounded corners, inner sep=\innersep, line width=\borderwidth] {
            \includegraphics[width=\imgwidth,height=\imgheight]{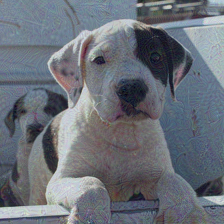}
        };
        &[\groupsep]
        \node[draw=blue!50, rounded corners, inner sep=\innersep, line width=\borderwidth] {
            \includegraphics[width=\imgwidth,height=\imgheight]{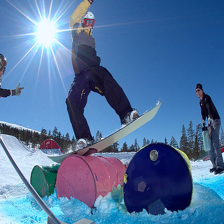}
        };
        &
        \node[draw=red!50, rounded corners, inner sep=\innersep, line width=\borderwidth] {
            \includegraphics[width=\imgwidth,height=\imgheight]{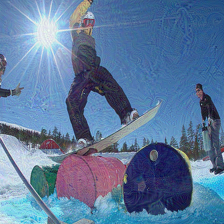}
        };
        \\
        
        \node[text width=\txtwidth, align=center, text=blue!50, font=\rmfamily\bfseries\small] {
            there are two dogs on the back of a boat
        };
        &
        \node[text width=\txtwidth, align=center, text=red!50, font=\rmfamily\bfseries\small] {
            a brown black and white dog and another black and white dog
        };
        &[\groupsep]
        \node[text width=\txtwidth, align=center, text=blue!50, font=\rmfamily\bfseries\small] {
           a man riding a snowboard on top of barrels
        };
        &
        \node[text width=\txtwidth, align=center, text=red!50, font=\rmfamily\bfseries\small] {
           surf boarder riding on the top of a wave
        };
        \\
    };
    \end{tikzpicture}

    \begin{tikzpicture}
    
    \matrix [column sep=0.01cm, row sep=0cm] {
        \node[draw=blue!50, rounded corners, inner sep=\innersep, line width=\borderwidth] {
            \includegraphics[width=\imgwidth,height=\imgheight]{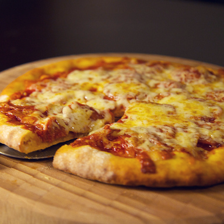}
        };
        &
        \node[draw=red!50, rounded corners, inner sep=\innersep, line width=\borderwidth] {
            \includegraphics[width=\imgwidth,height=\imgheight]{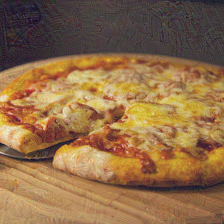}
        };
        &[\groupsep]
        \node[draw=blue!50, rounded corners, inner sep=\innersep, line width=\borderwidth] {
            \includegraphics[width=\imgwidth,height=\imgheight]{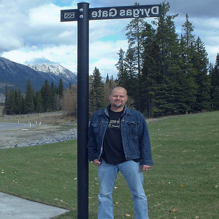}
        };
        &
        \node[draw=red!50, rounded corners, inner sep=\innersep, line width=\borderwidth] {
            \includegraphics[width=\imgwidth,height=\imgheight]{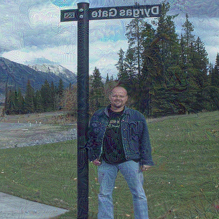}
        };
        \\
        
        \node[text width=\txtwidth, align=center, text=blue!50, font=\rmfamily\bfseries\small] {
            a small pizza on a cutting board with one slice displayed
        };
        &
        \node[text width=\txtwidth, align=center, text=red!50, font=\rmfamily\bfseries\small] {
            chicago style deep dish pizza with tomato sauce and sausage
        };
        &[\groupsep]
        \node[text width=\txtwidth, align=center, text=blue!50, font=\rmfamily\bfseries\small] {
            this is a man posing by a road sign that says dyrgas gate
        };
        &
        \node[text width=\txtwidth, align=center, text=red!50, font=\rmfamily\bfseries\small] {
            sign at the corner of clinton st and sw 68th st indicating salem exit approaching
        };
        \\
    };
    \end{tikzpicture}
    \caption{MS-COCO before and after distillation. \textcolor{blue}{(}\textit{\textcolor{blue}{Left}}\textcolor{blue}{)} The original image-text pairs before the distillation. \textcolor{red}{(}\textit{\textcolor{red}{Right}}\textcolor{red}{)} The 
     image-text pairs after distillation. }
    \label{fig:visualization_coco}
\end{figure}

Here we provide visualizations of distilled image-text pairs.  
\autoref{fig:visualization_flickr} and \autoref{fig:visualization_coco} present the original and distilled data on Flickr-30K and MS-COCO.  
The displayed texts are the closest matching sentences from the training set to the distilled text embeddings, following~\cite{wu2024visionlanguage}.

\end{document}